# Compact prior-residual neural networks generalize OCO-2 $XCO_2$ retrievals across viewing modes and five unseen years

Xianlei San[1], Chengxing Zhai[1], Hui Su[1], Kuai Dai[1], and Shuangchen Du[1]

[1]The Hong Kong University of Science and Technology, Clear Water Bay, Kowloon, Hong Kong SAR, China

Correspondence to: Chengxing Zhai (cxzhai@ust.hk)

**Abstract.** Full-physics retrieval of column-averaged dry-air $CO_2$ mole fraction ($XCO_2$) from Orbiting Carbon Observatory-2 spectra is costly, whereas machine-learning alternatives mostly target land-nadir soundings, in-period splits, or multi-million-parameter encoders. We pair bandwise singular value decomposition (30 scores per band) and nine ancillary and prior inputs with compact prior-residual multilayer perceptrons (MLPs; 16,893–164,793 parameters) to reproduce Atmospheric $CO_2$ Observations from Space (ACOS) Level-2 $XCO_2$ across global ocean-glint, land-glint, and nadir domains. Trained on 2015–2019 observations and validated on 2020, frozen MLPs generalize across 2021–2025 with mean annual root-mean-square errors (RMSEs) against ACOS of 0.89 (ocean glint), 1.35 (land glint), and 1.43 ppm (nadir), while CPU batches from in-memory spectra take 75.6–102.9 µs per sounding. Without regional retraining, the frozen global nadir MLP outperforms the ~6.5-million-parameter spectral Transformer over East Asia in 2020–2022 under aligned screening (1.376 vs. 1.503 ppm), and quarterly rolling updates cut 2021–2024 mean quarterly RMSE to 1.06 ppm. These results show that lightweight networks on linearly compressed spectra can rapidly emulate full-physics $XCO_2$ retrievals across viewing modes and multiple unseen years without heavy encoders while buffering random prior errors; remaining post-2022 bias drift motivates periodic updates and independent ground-based validation.

## 1 Introduction

Atmospheric $CO_2$ observations are needed to constrain the distribution and evolution of carbon sources and sinks. Atmospheric concentrations integrate anthropogenic emissions, terrestrial and oceanic carbon exchange, and transport, so their spatial and temporal variations provide complementary constraints on the global carbon budget (Friedlingstein et al., 2025). Spaceborne measurements of column-averaged dry-air $CO_2$ mole fraction ($XCO_2$) extend this information beyond the coverage of surface networks, provided that small atmospheric signals can be distinguished from variations in observing conditions (Eldering et al., 2017).

The Orbiting Carbon Observatory-2 (OCO-2) measures reflected sunlight in the $O_2$-A, weak-$CO_2$, and strong-$CO_2$ bands (Crisp et al., 2017). Operational full-physics retrievals estimate atmospheric and surface states by iteratively matching radiative-transfer simulations to these spectra (O'Dell et al., 2012, 2018; Taylor et al., 2026). This physically constrained approach accounts for absorption, multiple scattering by aerosols, and surface reflection, but repeated forward-model evaluations impose a substantial computational cost that has motivated growing interest in direct machine-learning retrieval and bias-correction approaches (Taylor et al., 2026).

Machine learning provides a fast alternative by mapping measured radiances directly to atmospheric $CO_2$. Early neural-network studies estimated tropical midtropospheric $CO_2$ from NOAA satellite observations (Chédin et al., 2003), whereas OCO-2 applications target full-column $XCO_2$. David et al. (2021) demonstrated empirical $XCO_2$ retrieval from land-nadir spectra using atmospheric-model estimates as training targets, and Bréon et al. (2022) extended this approach to land-glint observations, showing that excluding the weak-$CO_2$ band reduced date- and location-related spectral shortcuts and recovered emission-plume enhancements absent from the training targets. Both studies split contemporary observations into training and test subsets within the same multi-year period, using alternating even and odd months (David et al., 2021) or random partitions (Bréon et al., 2022), leaving multi-year transfer to future observation years untested.

Predicting $XCO_2$ in subsequent years is challenging because rising atmospheric $CO_2$ and changing observing conditions shift observations beyond the historical training distribution. Xie et al. (2024) augmented East Asian training observations with simulated spectra spanning higher $CO_2$ concentrations, whereas Chen et al. (2025) used a spectral Transformer (SpT) to predict the residual relative to the operational $XCO_2$ prior with periodic fine-tuning. A remaining question is how much model capacity spectral retrieval requires. Although OCO-2 spectra contain thousands of channels, correlated absorption lines concentrate most spectral variance into a low-dimensional subspace. Because large spectral encoders increase computational cost and the risk of overfitting to training-period conditions, we test whether linear spectral compression preserves the information needed for accurate retrieval with a compact network.

Surface reflectance and viewing geometry introduce additional radiative variability across OCO-2 observation modes. OCO-2 collects nadir (ND) and glint (GL) soundings along orbital tracks that cross both land and ocean (Crisp et al., 2017). Over water, directional reflection near the solar glint spot provides a strong shortwave-infrared signal, whereas diffuse nadir reflection is weak (Cox and Munk, 1954; O'Dell et al., 2018). Over land, heterogeneous surface reflectance, topography, and aerosols complicate radiative transfer in both GL and ND viewing geometries (Boesch et al., 2011; Massie et al., 2021). Evaluating GL ocean, GL land, and ND separately therefore clarifies how retrieval accuracy and multi-year error drift depend on the observation regime. Most published OCO-2 machine-learning studies have focused on land nadir observations (David et al., 2021; Xie et al., 2024; Chen et al., 2025). Land glint has also been investigated (Bréon et al., 2022), whereas ocean-glint evaluation remains comparatively limited.

Here we evaluate whether bandwise singular value decomposition (SVD) spectral compression, combined with prior and viewing-geometry predictors, enables compact prior-residual multilayer perceptrons (MLPs) to reproduce Atmospheric $CO_2$ Observations from Space (ACOS) Level-2 $XCO_2$ retrievals across global ocean-glint, land-glint, and nadir domains. Following Chen et al. (2025), the models predict corrections to the operational $XCO_2$ prior rather than absolute concentrations. Trained strictly on 2015–2019 observations and validated on 2020, the frozen networks are tested across five subsequent unseen years (2021–2025) and benchmarked against direct-$XCO_2$ targets, LightGBM models, and the published East Asian SpT results under aligned test-screening criteria. Through spectral-ablation, prior-perturbation, and bias-drift diagnostics, we quantify how leading spectral modes, prior sensitivity, and observing-system changes govern multi-year retrieval root-mean-square error (RMSE) and bias.

## 2 Data and methods

### 2.1 OCO-2 data and observation domains

We matched calibrated OCO-2 Level-1B spectra in the $O_2$-A (0.765 μm), weak-$CO_2$ (1.61 μm), and strong-$CO_2$ (2.06 μm) bands to ACOS Level-2 Standard records (builds B11006r–B11213r) using sounding identifiers. The Level-2 product supplies the retrieved $XCO_2$ reference (/RetrievalResults/xco2 in ppm), the operational $XCO_2$ prior (xco2_apriori), solar and viewing geometry, and surface-pressure priors (Crisp et al., 2017; O'Dell et al., 2018; Taylor et al., 2026). The $O_2$-A band constrains the dry-air column and effective optical path length, while the two $CO_2$ bands provide complementary absorption sensitivity (O'Dell et al., 2012; Bréon et al., 2022). Quality screening retained outcome_flag = 1, cloud_flag_abp = 0, and cloud_flag_idp ∈ {2, 3}, selecting clear A-band preprocessor cases and clear or probably clear IDP cases (Taylor et al., 2016). This screening reduces cloud-contaminated observations and excludes unsuccessful reference retrievals, limiting their influence on the learned spectral–$XCO_2$ relationship.

Screened soundings were assigned to three operational observation domains using viewing mode and the Level-2 wind-speed field. ND included all screened soundings with mode = ND. Within mode = GL, a finite nonnegative wind speed defined GL ocean and the fill value −999999 defined GL land; wind speed was used solely for domain assignment. Figure 1 shows the 2020 validation cohorts. Deterministic sampling by sounding identifier targeted 200,000 soundings per domain and year, followed by spectral-validity and duplicate screening and equal-domain rebalancing (Sect. S4). The final cohorts contained 1,000,000 training soundings per domain in 2015–2019 and 200,000 validation soundings in 2020 (Table S1). Holding out the next complete year makes model selection sensitive to post-training transfer. In the 2025 archive, GL land and ND extend through 30 September and GL ocean through 1 October (denoted 2025* in figures).

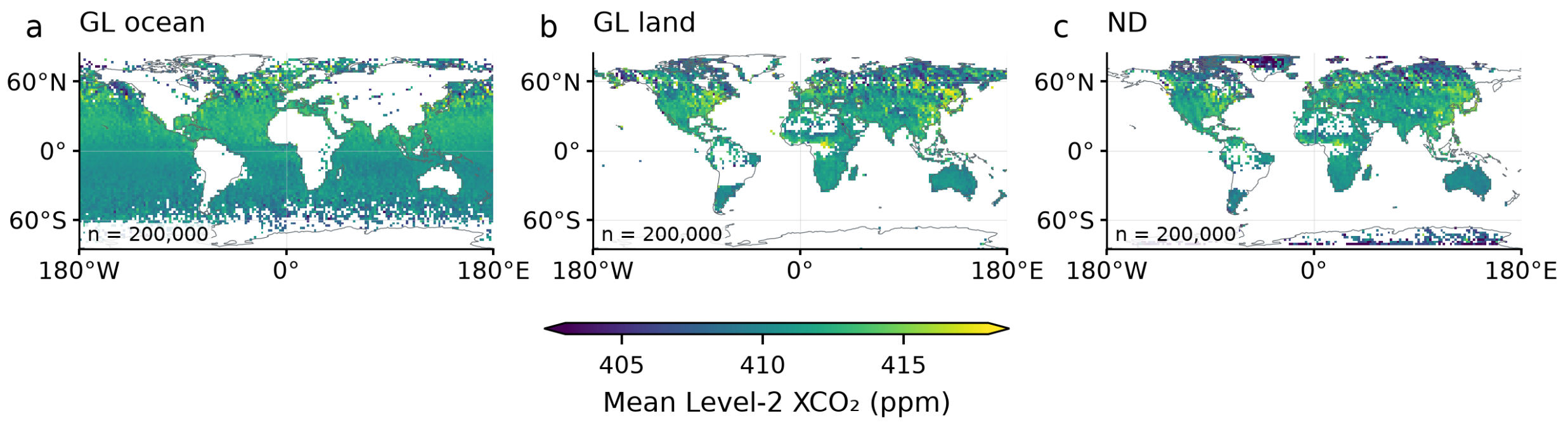

**Figure 1. Spatial distributions of annual mean ACOS Level-2 $XCO_2$ in 2° × 2° grid cells for the 2020 validation cohorts (200,000 soundings per domain): (a) GL ocean, (b) GL land, and (c) ND. Unsampled cells are left blank, and coastlines follow Natural Earth.**

### 2.2 Spectral preprocessing and SVD compression

All preprocessing parameters were determined from the 2015–2019 training archive. Because wavelength sampling is not exactly the same across detector footprints and observation dates, spectra were linearly resampled onto a common grid before SVD, ensuring that each matrix column represented the same wavelength across soundings. The reference grid used the mean dispersion relation from 173,672 valid training granule–footprint combinations. Degraded band edges were excluded, retaining the central 860, 925, and 881 channels in $O_2$-A, weak $CO_2$, and strong $CO_2$, respectively (Crisp et al., 2017; Xie et al., 2024) (Fig. 2). Each spectrum was normalized by the mean radiance of valid channels within a fixed 15-channel continuum window, reducing broadband amplitude differences while preserving relative absorption structure. The exact windows and trimming rule are documented in Sect. S1 and Fig. S4.

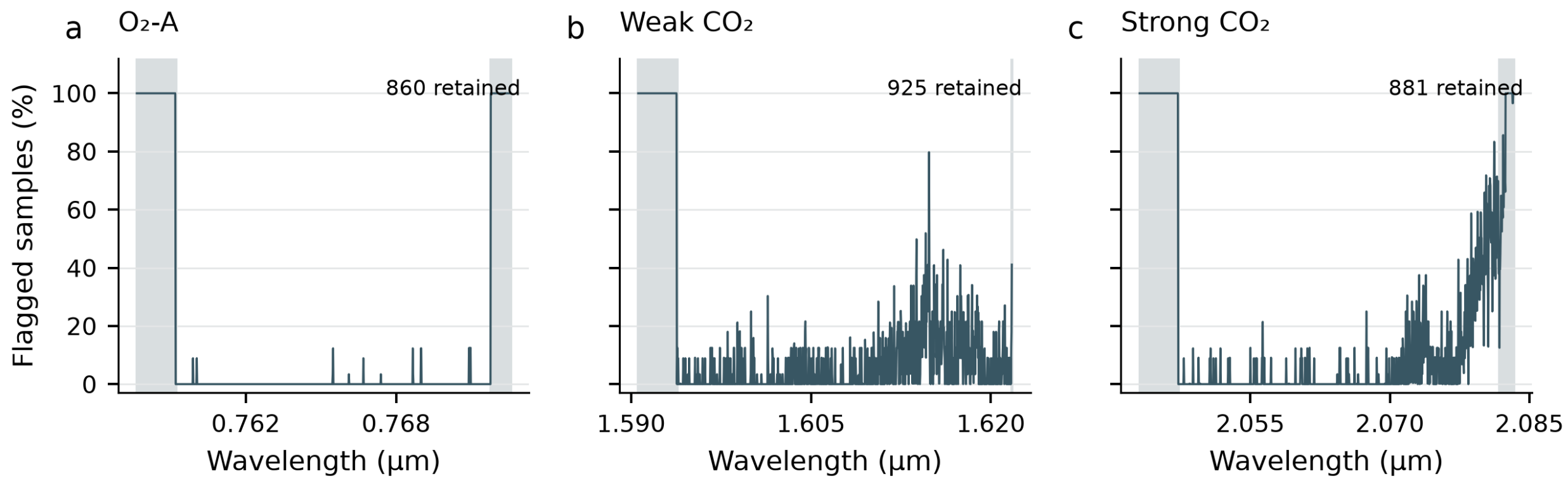

**Figure 2. Channel-wise bad-sample frequency across the 2015–2019 training archive (173,672 granule-footprints) for the (a) $O_2$-A, (b) weak-$CO_2$, and (c) strong-$CO_2$ bands. Shaded gray regions mark degraded band edges trimmed during preprocessing, while isolated interior bad pixels are masked during per-sounding SVD projection.**

For a target wavelength $\lambda$ between adjacent native detector channels $\lambda_j$ and $\lambda_{j+1}$, the resampled radiance $\tilde{L}_i(\lambda)$ is linearly interpolated, and the validity indicator $\tilde{m}_i(\lambda)$ equals unity only when both bracketing channels are unflagged:

$$\tilde{L}(\lambda) = (1 - \alpha)L_j + \alpha L_{j+1}, \quad \alpha = \frac{\lambda - \lambda_j}{\lambda_{j+1} - \lambda_j} \tag{1}$$

Each resampled spectrum was divided by the mean radiance of its valid channels within the fixed continuum window (Sect. S1; Fig. S4). A resampled channel was masked before coefficient estimation whenever either supporting native detector pixel was flagged.

Normalized training spectra were decomposed separately for each observation domain and band using uncentered truncated SVD on the common wavelength grid (Halko et al., 2011). After continuum normalization removes broadband radiance amplitude, uncentered SVD retains the mean normalized absorption shape directly within the basis vectors. The first 30 right-singular vectors per band capture 98.65 %–99.85 % of centered training spectral variance (Sect. 3.1):

$$X \approx U_k \Sigma_k V_k^{T}, \quad k = 30. \tag{2}$$

Here X is the matrix of normalized training spectra, and $V_k$ contains the first 30 basis vectors as columns. For sounding i, $G_i$ denotes its valid channel indices. The restricted basis $A_i = V_k[G_i, :]$ has $|G_i|$ rows and 30 columns, and $b_i = x_i[G_i]$ is the corresponding normalized spectrum as a column vector. Removing a flagged channel removes both its spectral value and the corresponding basis row. The 30 coefficients minimize the sum of squared reconstruction errors over valid channels and are obtained using the Moore–Penrose pseudoinverse $A_i^+$ (Penrose, 1955, 1956):

$$z_i = {A_i}^+ b_i \, . \tag{3}$$

The implementation verifies that more than 30 valid channels remain and that $A_i$ has full column rank before computing its pseudoinverse, yielding the unique least-squares solution while accounting for the non-orthogonality of the restricted basis columns after channel removal. Applying this masked least-squares projection across training, validation, and test soundings produces 30 coefficients per band and 90 spectral features in total.

### 2.3 Model inputs

The 90 spectral coefficients were combined with nine pre-retrieval ancillary and prior predictors to form the 99-input vector (Table 1). Solar zenith angle (SZA) and viewing zenith angle (VZA) were represented by their secants, encoding the geometric optical path scaling under a plane-parallel approximation where the air-mass factor scales as sec(SZA) + sec(VZA), reducing to sec(SZA) + 1 at nadir (Frankenberg et al., 2015). Relative azimuth angle (AZA) was represented by cos(AZA) and sin(AZA) to preserve angular periodicity, while Earth–Sun distance and relative solar velocity captured illumination and Doppler-shift geometry. Prior state inputs comprised the full-physics and A-band preprocessor surface-pressure priors (surface_pressure_apriori_fph and surface_pressure_apriori_abp) together with the operational $XCO_2$ prior (xco2_apriori) (O'Dell et al., 2018; Chen et al., 2025). All 99 inputs were standardized within each domain using the 2015–2019 training mean and standard deviation.

**Table 1. Summary of the 99 pre-retrieval input features used across all retrieval models.**

| Input category | Variables | Count |
|---|---|---|
| Spectral SVD scores | 30 projection scores per band: $O_2$-A (0.76 μm), weak $CO_2$ (1.61 μm), strong $CO_2$ (2.06 μm) | 90 |
| Viewing geometry | sec(SZA), sec(VZA), cos(AZA), sin(AZA)<br>SZA: solar zenith angle; VZA: viewing zenith angle; AZA: relative azimuth angle | 4 |
| Solar ephemeris | Earth–Sun distance, relative solar velocity | 2 |
| Prior state variables | Surface-pressure prior (FPH), ABP pressure prior, $XCO_2$ prior (xco2_apriori) | 3 |
| Total inputs | 90 compressed spectral features + 9 physical geometry and prior predictors | 99 |

## 2.4 Prior-residual learning

For sounding i, the training target is the prior residual $r_i$ between the Level-2 $XCO_2$ reference $y_i$ and its operational prior $a_i$. Each regressor estimates $\hat{r}_i$ from the 99 inputs, and the retrieved concentration $\hat{y}_i$ is reconstructed by adding the predicted residual to the prior:

$$r_i = y_i - a_i, \quad \hat{y}_i = a_i + \hat{r}_i \tag{4}$$

Following Chen et al. (2025), the operational prior provides an observation-specific background and the regressor predicts the difference between the ACOS retrieval and that prior. The prior is retained as an input so that the correction can depend on its value. Adding the predicted residual back to the prior gives the $XCO_2$ estimate, with identical prediction errors in residual and $XCO_2$ space ($\hat{y}_i - y_i = \hat{r}_i - r_i$). We compare direct-$XCO_2$ and residual prediction under matched inputs and training settings to evaluate the effect of this target formulation (Sect. 3.2).

## 2.5 Regression models and evaluation

Within each domain, we compared validation-selected LightGBM models (Ke et al., 2017) with a fully connected MLP trained using AdamW (Loshchilov and Hutter, 2019) on the 99-input vector and prior-residual target (Fig. 3). With domain-specific MLP training hyperparameters fixed (Tables S7 and S12), a Tree-structured Parzen Estimator search evaluated 128 candidate architectures with one to four hidden layers and 32–640 units per layer. The top three completed architectures were trained with seeds 17, 29, and 43 and selected by mean 2020 validation RMSE (Sect. S2). The selected hidden-layer widths were 220 for GL ocean, 164 for GL land, and 468–248 for ND. Each MLP minimized minibatch mean squared error on standardized training residuals with gradient-norm clipping at 5 and domain-specific weight decay.

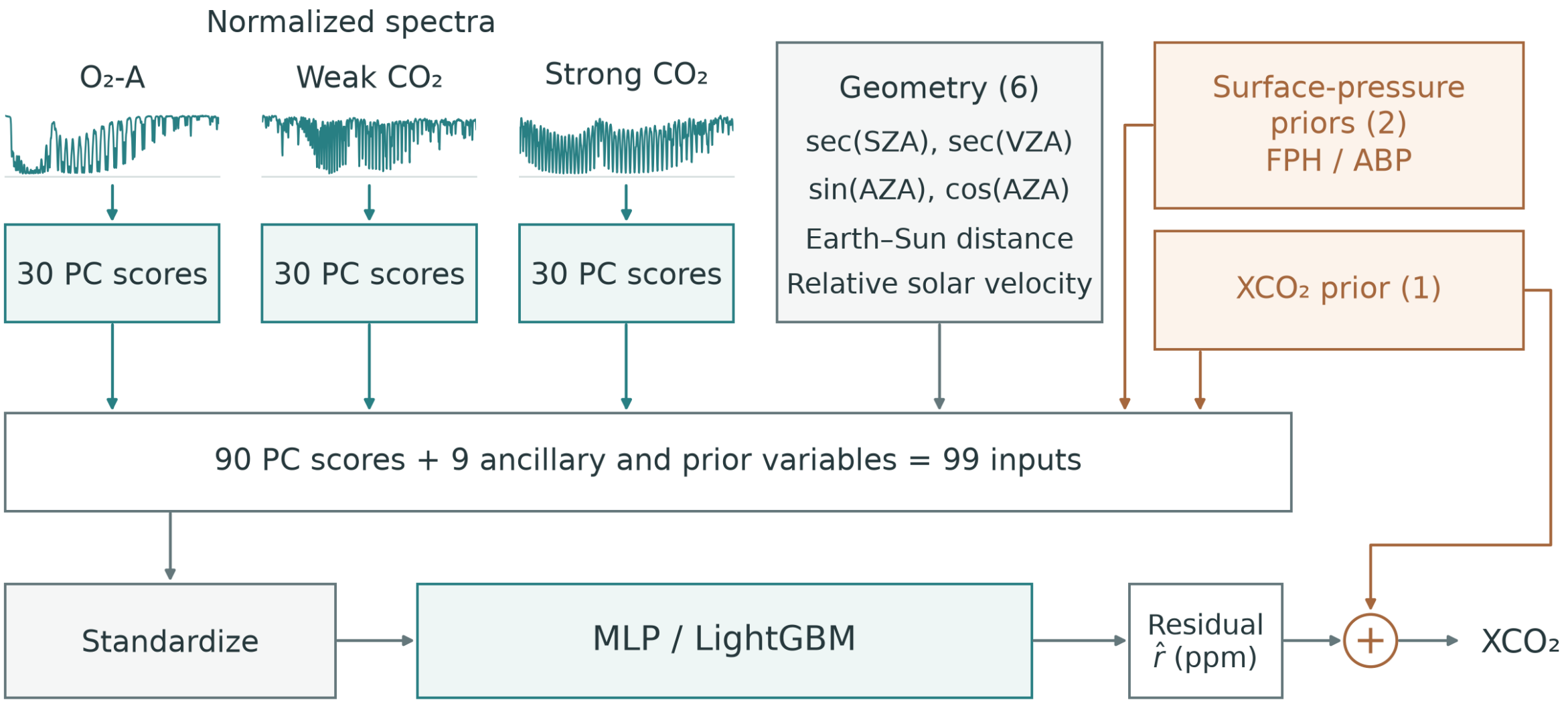

**Figure 3. Shared prior-referenced retrieval workflow for MLP and LightGBM. Each normalized band is compressed by SVD and masked least-squares projection to 30 spectral scores. The 90 scores are combined with six geometry and ephemeris variables, two surface-pressure priors (FPH and ABP), and the $XCO_2$ prior, then standardized using training-set statistics. Either regressor predicts the prior residual, which is converted to ppm and added to the $XCO_2$ prior to reconstruct $XCO_2$.**

LightGBM selection screened 24 predefined hyperparameter settings with seed 17, confirmed the top three settings and the default reference setting across seeds 17, 29, and 43, and selected the configuration with the lowest mean 2020 validation RMSE (Sect. S2). Predicted residuals were converted to ppm and added to the operational prior, and reconstructed-$XCO_2$ RMSE, mean bias (prediction minus reference), and $R^2$ were evaluated for each year and seed.

To evaluate the effect of the prediction target, we compared direct-$XCO_2$ and prior-residual regression using independently selected models for each target (Sect. 3.2). Within each model family and domain, both targets used the same 99 inputs, training and validation cohorts, random seeds, and hyperparameter search spaces described above (up to 500 training epochs with a 40-epoch minimum and 30-epoch early-stopping patience for the MLPs). All 12 selected models were frozen before evaluation on the 2021–2025 test cohorts.

### 2.6 East Asian benchmark using SpT screening criteria

To match the test-data screening of Chen et al. (2025), we constructed an East Asian ND test cohort from archived ACOS Level-2 records and matched OCO-2 Level-1B spectra following the public SpT preprocessing criteria: successful retrieval (outcome_flag = 1), 100 % land fraction, and spatial bounds of 20° N–45° N and 110° E–145° E across footprints 1–8, without the additional ABP and IDP cloud filters used in our global cohorts. After matching by sounding identifier, removing invalid spectra and duplicates, and excluding 5,479 soundings present in the 2020 global ND validation cohort, the regional test set comprised 475,134 soundings across 2020–2025 (Sect. S4; Table S1).

For the annual comparison, we applied the frozen global ND residual MLP directly to this regional cohort using its 2015–2019 SVD bases, training scalers, and three checkpoints selected on the 2020 global validation set. Over the common 2020–2022 test period, we compare annual RMSEs against ACOS and their three-year mean with the published base SpT results (Sect. 3.4).

The quarterly comparison starts from the same global ND checkpoints. For each target quarter from 2021 Q1 to 2024 Q1, we compare frozen weights with rolling fine-tuning on earlier observations. Both variants use identical target soundings from the SpT-screened East Asian cohort (Sect. 3.4; Sect. S5).

### 2.7 Spectral and prior sensitivity controls and error diagnostics

Spectral component controls used the same 1,000,000 training soundings per domain and 200,000-sounding 2020 validation cohorts as the main experiment. Each configuration combined the nine ancillary and prior predictors with 0, 10, 20, or 30 coefficients per band; the zero-coefficient configuration used only the nine non-spectral inputs. Lower-dimensional configurations took prefixes of the 30-component masked least-squares scores. Domain-specific hidden-layer architectures, batch sizes, and regularization settings were held fixed across component counts. Training allowed up to 500 epochs, with early stopping after at least 40 epochs and a patience of 30 epochs for a minimum validation-RMSE improvement of $10^{-5}$ ppm, evaluating the best validation checkpoint (Sect. S2). Checkpoints and learning rates were selected on the 2020 validation set and then evaluated across the 2021–2025 test cohorts (Sect. 3.6; Table S10).

We tested sensitivity to random $XCO_2$-prior errors without retraining the models. Each global domain retained its 2020 validation and 2024 test cohorts of 200,000 soundings per year. Zero-mean Gaussian perturbations with standard deviations of 0.5, 1, and 2 ppm were added simultaneously to the prior input feature and the prior used to reconstruct $XCO_2$ from the predicted residual, with the unperturbed prior serving as the baseline. Both the perturbed prior and the resulting model retrievals were evaluated against the unperturbed ACOS reference, with all other inputs and historical scalers held fixed across both model families and all three seeds (Sect. S6).

For each main-model sounding, prediction error was defined as reconstructed $XCO_2$ minus the ACOS reference. Annual mean bias and its 95 % confidence interval were estimated from the 2.5th and 97.5th percentiles of 1,000 within-year date-block bootstrap draws, resampling observation dates while retaining all soundings and three seeds on each drawn date (Sect. 3.6). Latitudinal bias profiles were computed across the complete years 2021–2024 by averaging errors over the three seeds and all soundings within 10° latitude bands containing at least 30 observations (Sect. 3.6).

## 3 Results

### 3.1 Compact spectral representation

Across the nine domain–band combinations, the first 30 SVD components captured 98.65 %–99.85 % of centered training spectral variance, leaving 0.15 %–1.35 % in higher-order components (Fig. 4). Residual variance declined steeply within the first 10 components for the $O_2$-A and weak-$CO_2$ bands (Fig. 4a, b) and within the first 20 components for the strong-$CO_2$ band (Fig. 4c), flattening thereafter.

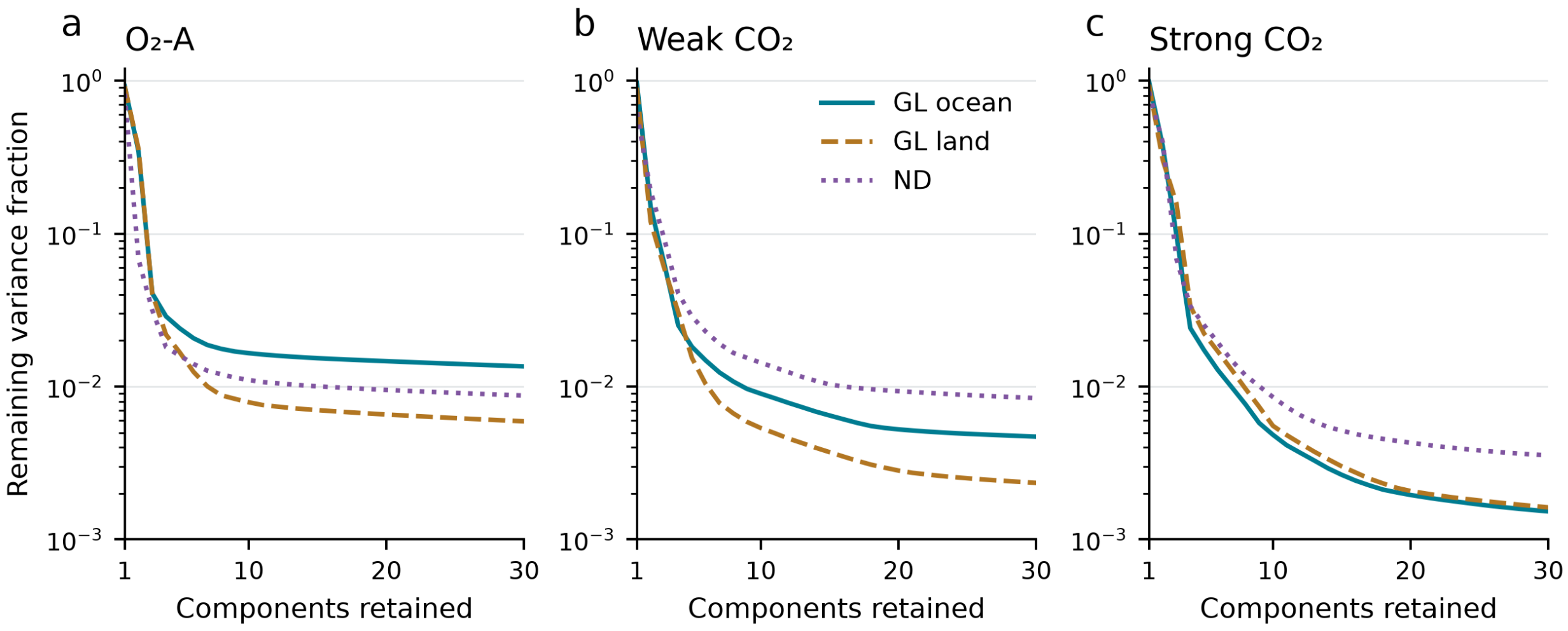

**Figure 4. Fraction of remaining centered spectral variance as a function of the number of retained SVD components (up to 30) for the (a) $O_2$-A, (b) weak-$CO_2$, and (c) strong-$CO_2$ bands across GL ocean, GL land, and ND on the 2015–2019 training set (1,000,000 spectra per domain).**

### 3.2 Residual variability and prediction-target comparison

Across GL ocean, GL land, and ND, annual mean ACOS $XCO_2$ increased by 24.8–25.7 ppm between 2015 and 2025, accompanied by similar growth in the operational prior (Fig. 5a–c). The annual mean reference-minus-prior residuals varied over much smaller ranges: −1.22 to −0.16 ppm for GL ocean, −1.56 to −0.43 ppm for GL land, and −2.04 to −0.62 ppm for ND (right axes in Fig. 5a–c). Mean residual offsets differed among domains. Subtracting the prior therefore removes most of the long-term background growth while retaining domain-dependent offsets and interannual variability for the regressors to capture.

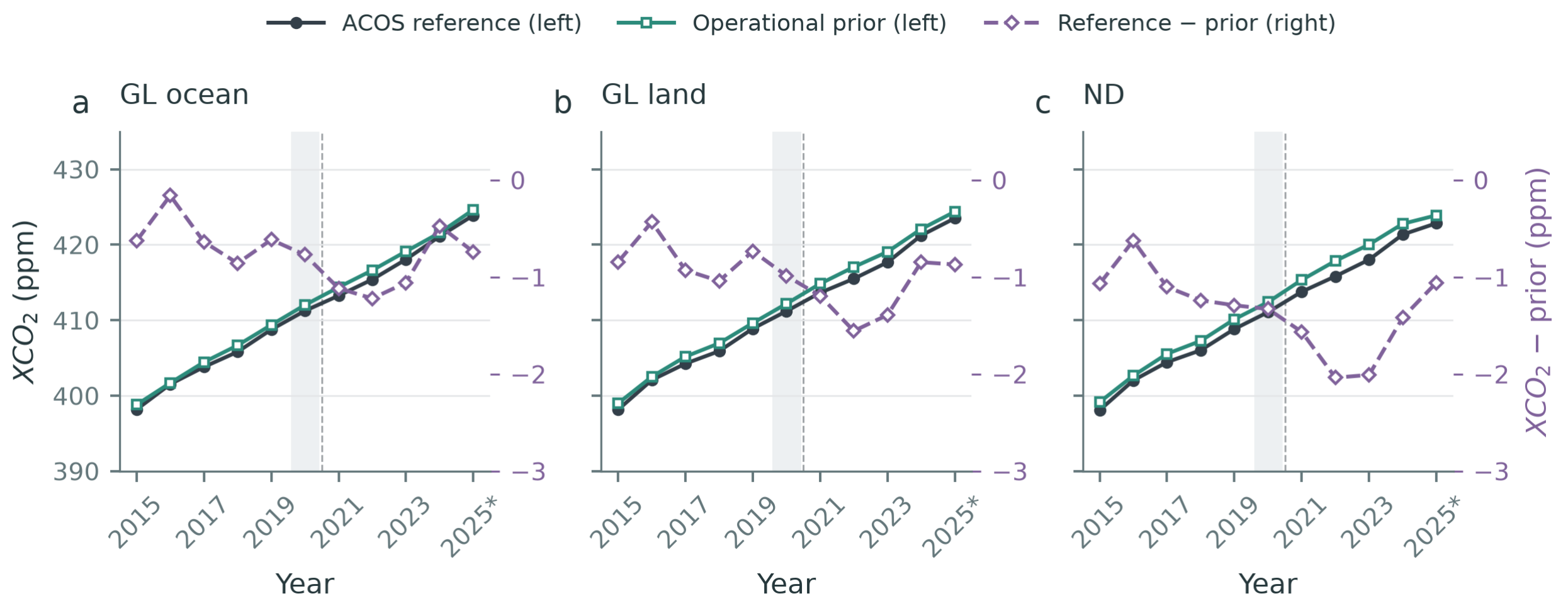

**Figure 5. Annual mean ACOS $XCO_2$, operational prior and their difference in 2015–2025 for (a) GL ocean, (b) GL land and (c) ND. Reference $XCO_2$ and the prior use the left axes; purple dashed curves show reference minus prior on the right axes. Each point is an annual sample mean. Left-axis limits are shared across panels, as are right-axis limits. Pale bands mark 2020 validation, and dashed vertical boundaries separate the later test years. Sample counts are given in Table S1.**

The separately optimized direct and residual target-control MLPs performed similarly on the 2020 validation set, where residual prediction reduced RMSE by 0.02–0.06 ppm across the three domains (Fig. 6). The difference grew across the 2021–2025 test period: mean annual MLP test RMSE decreased from 1.60 to 0.89 ppm over GL ocean, from 1.68 to 1.33 ppm over GL land, and from 1.51 to 1.43 ppm for ND. The reduction was larger for LightGBM, whose mean annual test RMSEs dropped from 6.49, 6.46, and 6.79 ppm under direct prediction to 1.42, 1.72, and 1.81 ppm under residual prediction. The RMSE advantage of residual prediction widened each year for the MLP over GL ocean and GL land and for LightGBM across all three domains, consistent with the narrower interannual range of the prior-residual target (Fig. 5). The land-glint residual MLP in this comparison was selected independently of the main land-glint MLP reported in Sect. 3.3.

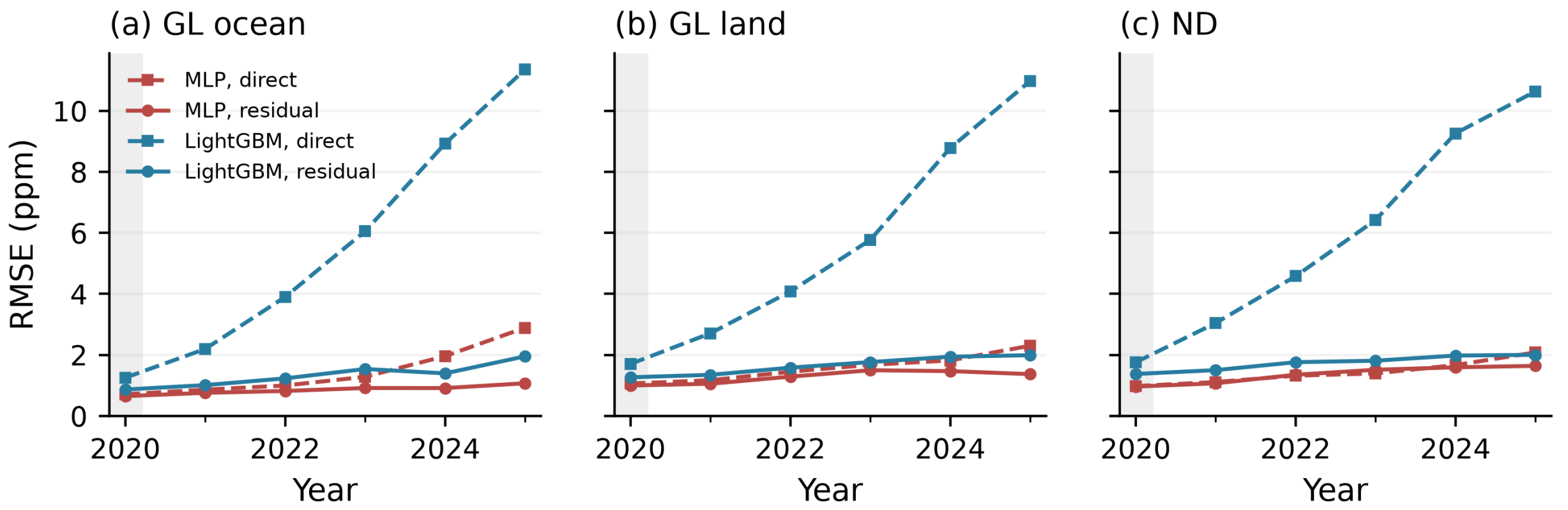

**Figure 6. Direct-XCO₂ and prior-residual prediction under independent validation selection. Annual RMSEs are evaluated against ACOS for validation in 2020 and testing in 2021–2025. Dashed lines with squares denote direct prediction, and solid lines with circles denote residual prediction; red indicates MLP and blue indicates LightGBM. Points show means across seeds 17, 29, and 43. Within each model family, both targets share identical cohorts, 99-feature inputs, and hyperparameter search spaces.**

### 3.3 Retrieval performance across observation domains

Across all three observation domains and every test year, the residual MLP improved upon the operational prior relative to ACOS and achieved lower RMSE than the validation-selected LightGBM models (Fig. 7; Table 2). Mean annual 2021–2025 MLP RMSEs were 0.89, 1.35, and 1.43 ppm for GL ocean, GL land, and ND, respectively, corresponding to 48.7 %, 49.4 %, and 57.9 % reductions relative to the uncorrected prior. Over GL ocean in 2025, LightGBM RMSE reached 1.95 ppm, exceeding the prior RMSE of 1.64 ppm, whereas the MLP maintained an RMSE of 1.07 ppm (Fig. 7a). We therefore focus subsequent analyses on the residual MLP. Even with residual learning, frozen-model test errors increased gradually over 2021–2025, most notably in ND (Fig. 7c).

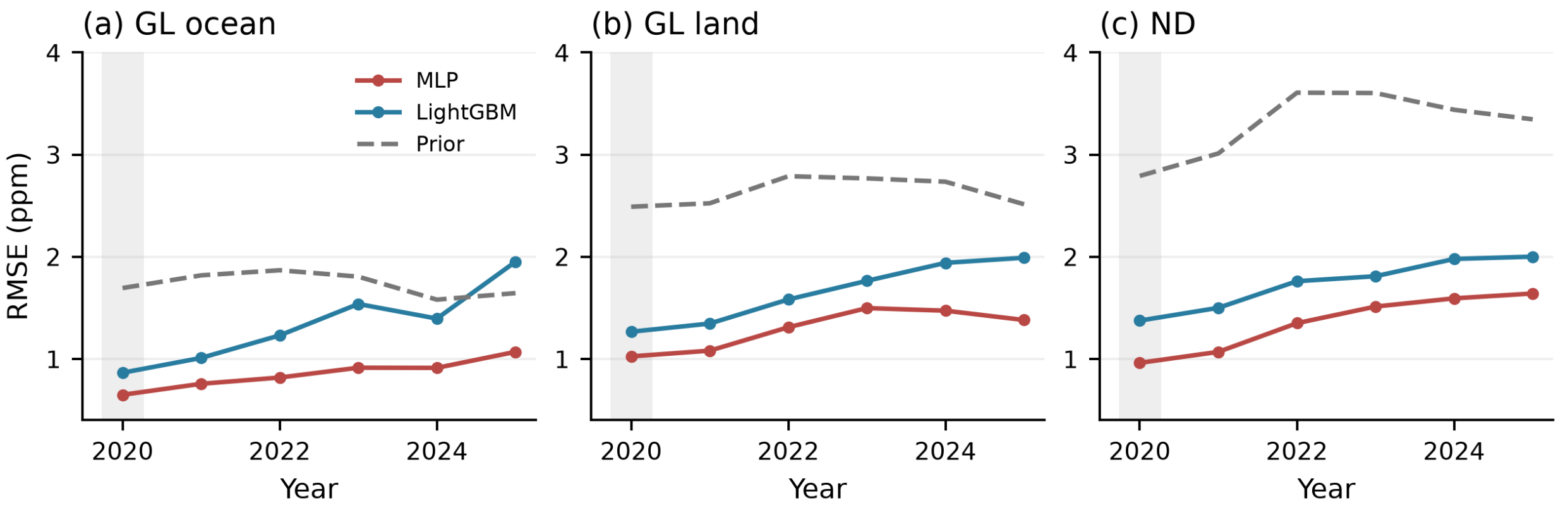

**Figure 7. Annual validation (2020) and test (2021–2025) RMSE against ACOS for (a) GL ocean, (b) GL land, and (c) global ND. Points and lines show the mean RMSE across seeds 17, 29, and 43. The gray dashed line shows the uncorrected operational prior. The pale band marks the validation year.**

Over GL land, annual MLP RMSE tracked the temporal shape of the prior RMSE, rising through 2023 and declining in 2024–2025 while remaining 1.14–1.47 ppm below the prior in every year (Fig. 7b). In ND, prior RMSE rose from 3.01 ppm in 2021 to 3.61 ppm in 2022 (Fig. 7c), driven by increases in both the prior mean bias (1.56 to 2.04 ppm) and error standard deviation (2.57 to 2.98 ppm). Over the same interval, MLP RMSE increased by 0.28 ppm (from 1.07 to 1.35 ppm), attenuating more than half of the 2022 jump in prior error.

**Table 2. Main-cohort performance across domains.**

| Metric | GL ocean | GL land | ND |
|---|---|---|---|
| Prior RMSE | 1.743 | 2.665 | 3.401 |
| Prior $R^2$ | 0.345 | 0.179 | 0.179 |
| MLP RMSE | 0.894 | 1.348 | 1.433 |
| MLP $R^2$ | 0.829 | 0.789 | 0.855 |
| LightGBM RMSE | 1.424 | 1.724 | 1.810 |
| LightGBM $R^2$ | 0.558 | 0.652 | 0.769 |
| RMSE reduction | 48.7 % | 49.4 % | 57.9 % |
| 2025 MLP bias | +0.452 | +0.632 | +0.651 |

**Note: RMSE and $R^2$ are unweighted means of annual metrics for 2021–2025; model metrics are first averaged across three random seeds. Bias refers only to the MLP in 2025. RMSE and bias are in ppm; reduction is the percentage decrease in mean annual MLP RMSE relative to the operational prior.**

### 3.4 East Asian nadir results under SpT screening criteria

Without regional retraining, the global ND MLP achieved annual RMSEs against ACOS of 1.287, 1.351 and 1.488 ppm in 2020–2022, respectively. These were lower in all three years than the corresponding published base SpT values (1.316, 1.526 and 1.666 ppm). The mean annual RMSE was 1.376 ppm, 8.4 % lower than the published SpT mean of 1.503 ppm (Fig. 8a).

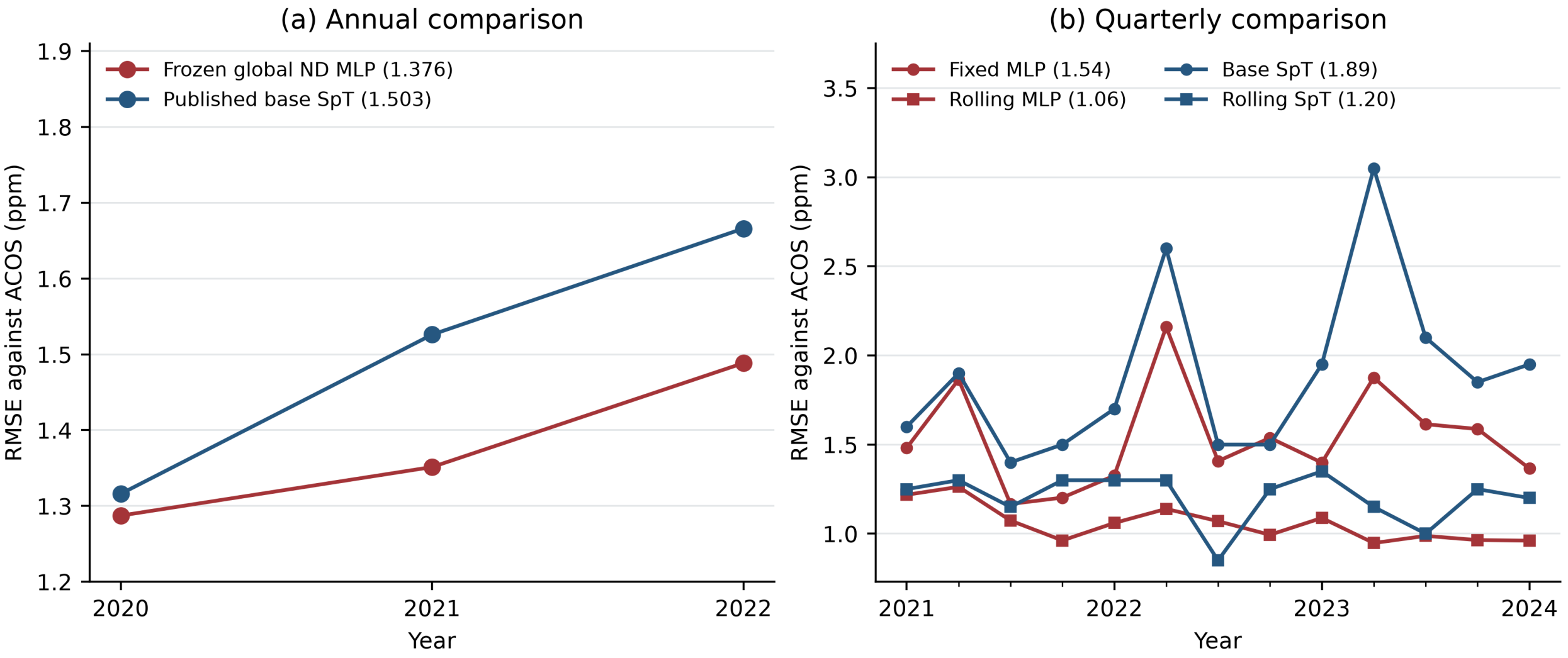

**Figure 8. East Asian ND RMSE against ACOS. (a) Frozen global ND MLP evaluated on the SpT-screened annual cohort and published base SpT, 2020–2022. (b) Fixed and rolling global ND MLPs on the SpT-screened quarterly cohort and published SpT series, 2021 Q1–2024 Q1. Red denotes MLP, blue denotes SpT, circles denote fixed weights and squares denote rolling updates. Legends give equal-year or equal-quarter means, respectively. MLP values are three-seed means.**

When the same frozen model was evaluated through 2025, the annual East Asian RMSEs for 2023–2025 were 1.537, 1.476 and 1.546 ppm (Table S6).

Across 13 quarters, rolling updates reduced the global ND MLP's mean quarterly RMSE against ACOS from 1.54 to 1.06 ppm. The published base and rolling SpT means were approximately 1.89 and 1.20 ppm, respectively (Fig. 8b). The rolling MLP also showed less quarter-to-quarter variation than the published rolling SpT series (standard deviations of 0.10 and approximately 0.14 ppm). The rolling models had relatively similar mean quarterly RMSEs in 2021 (1.13 ppm for the MLP versus approximately 1.25 ppm for SpT). The gap was larger in 2023 (1.00 versus approximately 1.19 ppm) and in the first quarter of 2024 (0.96 versus approximately 1.20 ppm).

### 3.5 Measured retrieval computation time

We benchmarked two stages separately on an Intel Xeon Platinum 8581C CPU (four threads) and an NVIDIA RTX 5880 Ada Generation GPU. Starting from 1,024 native Level-1B spectra and associated inputs already in memory, the residual MLP pipeline took 75.6–102.9 µs per sounding on the CPU across observation domains (Table S9). For GL ocean, the batch took 0.105 s (102.9 µs per sounding). The median latency for a single GL-ocean sounding was 1.33 ms. This benchmark includes wavelength registration, continuum normalization, masked least-squares projection, feature assembly and scaling, and prediction with the prior added back. It excludes disk I/O, cloud screening, prior generation, and model loading.

Starting instead from 200,000 precomputed 99-feature inputs, the validation-selected LightGBM models took 26.683, 29.364, and 28.084 s on the four-thread CPU for GL ocean, GL land, and ND, respectively. The corresponding MLP times were 0.193, 0.202, and 0.409 s (Table S8). For GL ocean, the MLP took 0.96 µs per sounding in the CPU batch and 0.137 s for the full GPU batch. Starting from native spectra in memory, LightGBM took 227.8–234.7 µs per sounding in 1,024-sounding CPU batches (Table S9). For context, Chen et al. (2025) reported more than 2–3 min per sounding for full-physics retrieval and 6.2 ms per sounding for SpT on an Intel Xeon Gold 5218R CPU. Our global ND MLP took 1.20 ms for a single sounding starting from in-memory native spectra on the Intel Xeon Platinum 8581C. These published times provide context; differences in hardware and processing scope preclude a direct speedup estimate.

### 3.6 Spectral and prior contributions and temporal error diagnostics

Including spectral scores lowered validation and multi-year test RMSE relative to the nine ancillary and prior predictors alone across all three domains (Fig. 9). For GL ocean and GL land, RMSE decreased steeply up to 20 scores per band, with smaller reductions from 20 to 30 scores. In ND, the largest reduction occurred within the first 10 scores per band, followed by a nearly flat response above 10 scores.

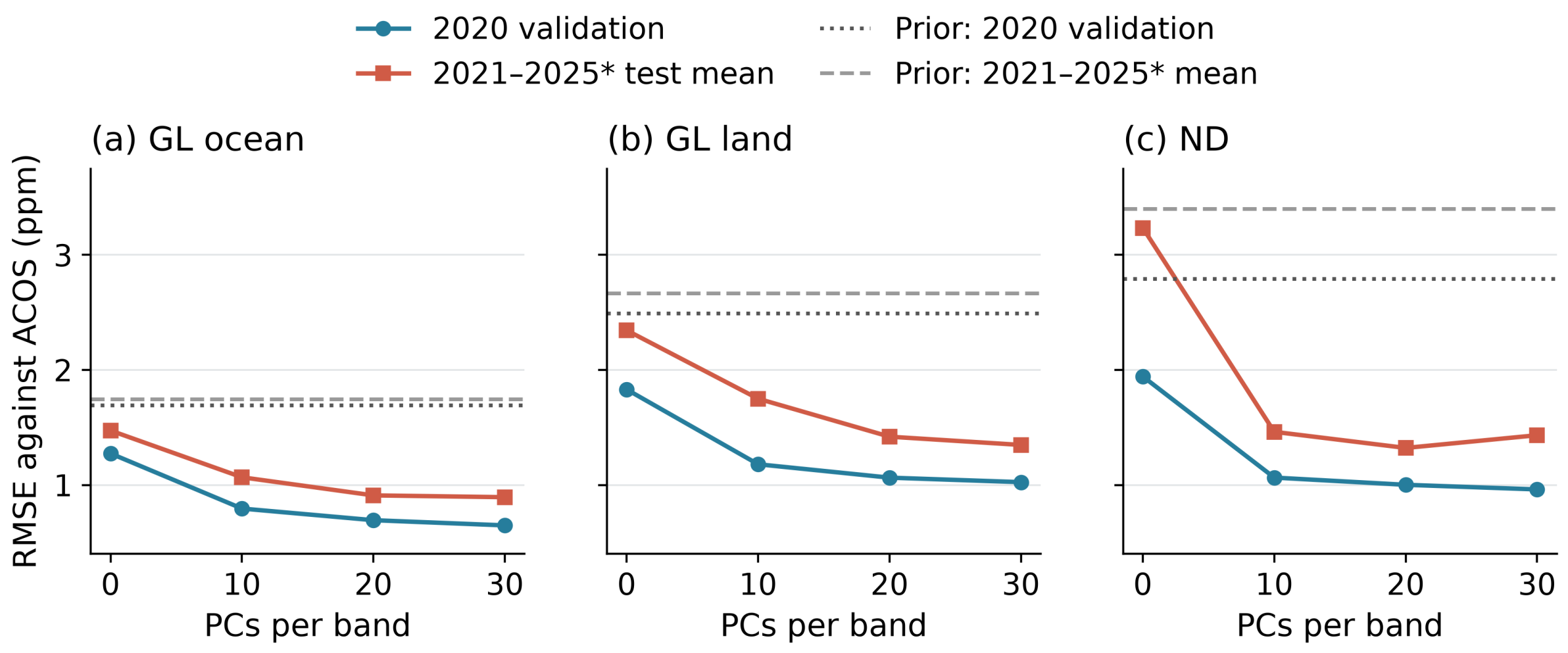

**Figure 9. Spectral component sensitivity with prior-error baselines: (a) GL ocean, (b) GL land and (c) ND. Each domain uses 1,000,000 training and 200,000 validation soundings. All configurations retain nine ancillary and prior predictors plus 0, 10, 20 or 30 coefficients per band. Points and lines show three-seed mean RMSEs. Horizontal dotted and dashed lines show the uncorrected prior RMSE for 2020 validation and the 2021–2025 mean, respectively, on the same annual cohorts. Test RMSEs are averaged equally across years within each seed before seed aggregation. Model recipes are fixed across component counts, with the common GL-land cosine horizon specified in Sect. S2.**

In global ND, increasing from 20 to 30 scores per band lowered 2020 validation RMSE while increasing mean test RMSE across 2021–2025, indicating that fitting fine-scale variations in the validation year can trade off against multi-year transfer stability. The zero-PC models also improved upon the uncorrected prior in post-training years, showing that ancillary and prior predictors contribute baseline state adjustments while the spectral scores drive the primary accuracy gains.

Adding zero-mean Gaussian noise with a standard deviation of 1 ppm to the 2024 priors increased prior RMSE from 1.580, 2.734 and 3.438 ppm to 1.871, 2.913 and 3.579 ppm for GL ocean, GL land and ND, respectively. The corresponding MLP RMSEs increased from 0.912, 1.472 and 1.593 ppm to 1.075, 1.566 and 1.670 ppm. The smaller increase in MLP RMSE than in prior RMSE across all three domains indicates moderate MLP sensitivity to the $XCO_2$ prior under the tested random perturbations (Fig. 10).

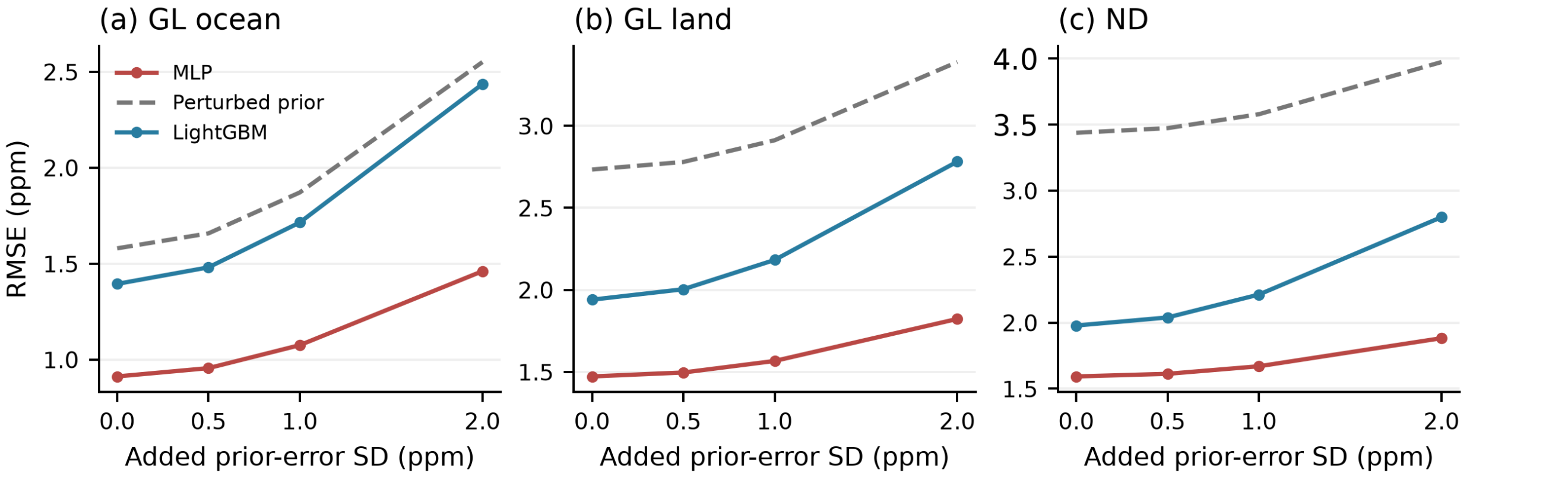

**Figure 10. Sensitivity to random $XCO_2$-prior errors in the complete 2024 test cohorts: (a) GL ocean, (b) GL land and (c) ND. The horizontal axis gives the standard deviation of the added zero-mean Gaussian error. The prior input and the prior added to the predicted residual are perturbed together. Model curves show three-seed mean RMSE, and the gray dashed line shows perturbed-prior RMSE; both are evaluated against the same ACOS reference. Models share the same perturbation vector and 200,000 soundings per domain. Zero noise denotes the original baseline.**

Main-model annual mean biases were positive relative to ACOS across 2021–2025 and varied by observation domain (Fig. 11). Bias over GL ocean remained comparatively stable and decreased in 2024, whereas GL land and ND exhibited larger positive shifts after 2022. In the 2025 cohort, mean biases (with 95 % date-block bootstrap intervals) were +0.452 ppm (+0.415 to +0.489) for GL ocean, +0.632 ppm (+0.585 to +0.682) for GL land, and +0.651 ppm (+0.564 to +0.727) for ND. Latitude profiles for 2021–2024 show comparatively consistent patterns over GL ocean (Fig. 12). GL land exhibits southern high-latitude underestimation in several years and stronger northern positive biases from 2022 onward. ND profiles change more substantially with latitude and year, without a similarly persistent pattern.

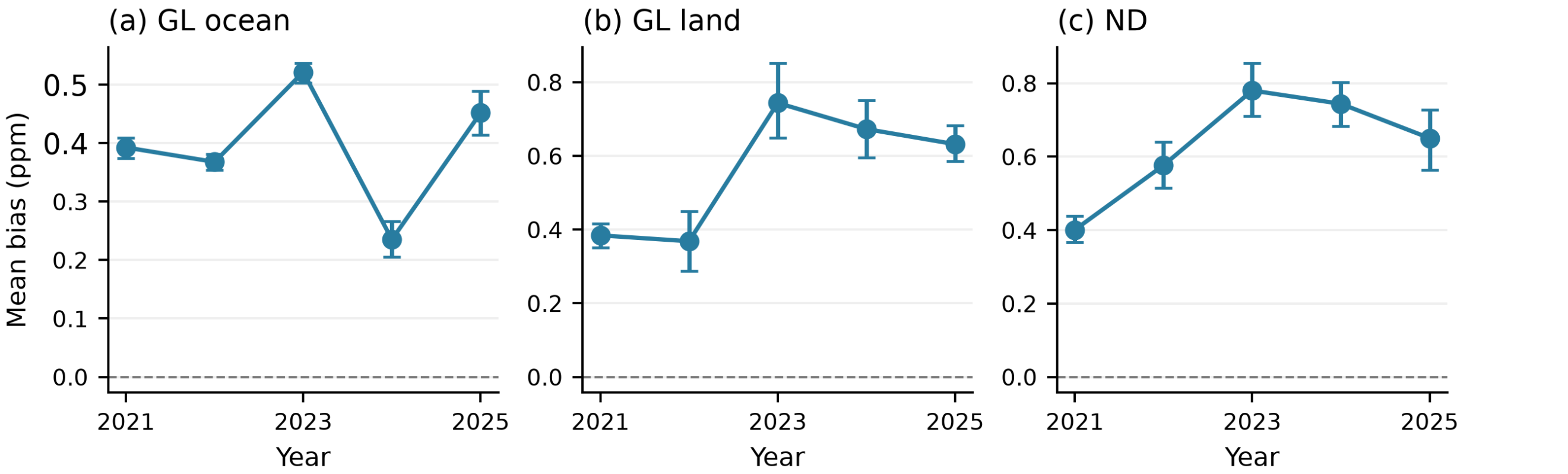

**Figure 11. Annual mean prediction-minus-ACOS bias of the main MLPs in 2021–2025: (a) GL ocean, (b) GL land, and (c) ND. Points show the three-seed mean bias; vertical error bars indicate 95 % intervals from 1,000 within-year date-block bootstrap draws with same-day soundings and seeds paired. The dashed line marks zero bias.**

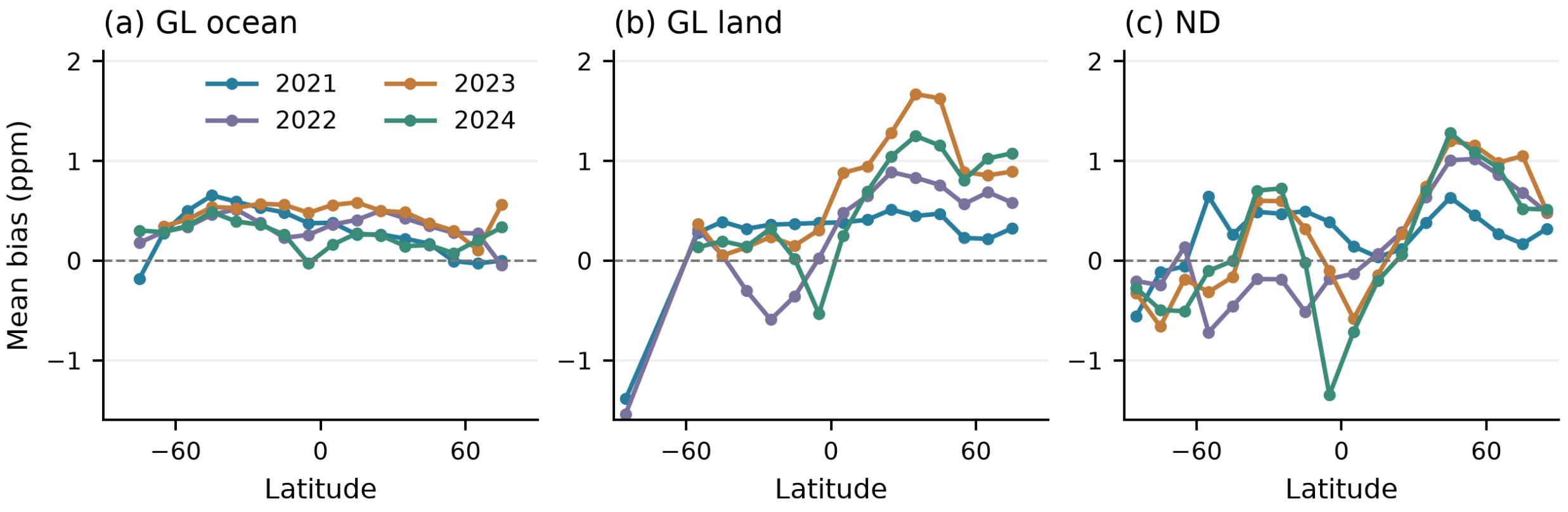

**Figure 12. Latitude-dependent mean bias of the main MLPs relative to ACOS during the complete years 2021–2024: (a) GL ocean, (b) GL land, and (c) ND. Each point averages prediction-minus-reference errors across three seeds and all soundings within a 10° latitude band containing at least 30 observations. The horizontal dashed line marks zero bias.**

## 4 Discussion

### 4.1 Spectral information and prior correction

The spectral component controls separate baseline prior adjustments from spectral retrieval skill (Fig. 9). Without spectral inputs, the zero-score model learns a conditional correction from the $XCO_2$ prior, surface-pressure priors, and viewing geometry, reflecting systematic relationships between the ACOS retrieval and its prior state. Including the leading SVD components produces the primary accuracy gains, which persist across the 2021–2025 test period. Because SVD orders basis vectors by spectral variance rather than predictive relevance to $XCO_2$ (Jolliffe, 1982; Halko et al., 2011), higher-order components mix weak absorption signals with measurement noise and fine-scale surface or instrument variability. In global ND, increasing from 20 to 30 components per band lowered 2020 validation RMSE but slightly increased 2021–2025 test RMSE, indicating that higher-order nadir components capture validation-year variations that transfer less reliably to subsequent years.

In the prior-perturbation experiments, adding random noise increased MLP RMSE by less than the added prior error in all three domains (Fig. 10). Because the operational prior enters both the MLP input vector and the final reconstruction in Eq. (4), the network uses spectral and geometric inputs to adjust the predicted residual when the prior departs from the state implied by the radiance spectrum, partially compensating for random prior errors.

The remaining interannual error drift reflects shifts in both the prior residuals and the observing system. ACOS v11 $CO_2$ priors represent large-scale background growth and climatological profile structure derived from NOAA marine boundary-layer measurements at Mauna Loa and American Samoa with approximately one-month latency (Laughner et al., 2023; Jacobs et al., 2024; Taylor et al., 2026). Subtracting the prior removes the global secular trend (Fig. 5), whereas regional fluxes, seasonal cycles, atmospheric transport, instrument calibration drift, and operational updates continue to evolve across years. Taylor et al. (2026) documented several observing-system and prior changes during our 2023–2025 test period, including the April 2024 transition from GEOS FP-IT to GEOS IT meteorological priors, operational $CO_2$-prior interpolation errors beginning in February 2025 after an American Samoa station data outage, and time-dependent L1B radiometric drift from detector ice build-up during the two-year February 2023–February 2025 interval without a decontamination cycle. Such shifts in measured radiances and operational priors alter the residual distribution relative to the 2015–2019 training period, consistent with the interannual residual variations in Fig. 5 and the domain-dependent bias trends in Fig. 11.

Spatial and temporal bias diagnostics identify the regimes that drive these post-2022 shifts (Figs. 11 and 12). Over GL ocean, latitudinal bias profiles remained within ±0.7 ppm across 2021–2024 (Fig. 12a). Over GL land, negative biases of −1.39 and −1.54 ppm occurred at 80°–90° S in 2021 and 2022, where the test cohorts contained only 57 and 69 soundings over high-albedo snow and ice surfaces, while positive biases strengthened across northern midlatitudes after 2022 (Fig. 12b). In ND, latitudinal biases varied more strongly from year to year across the tropics and northern midlatitudes (Fig. 12c).

### 4.2 Interpreting the ocean-glint result

Among the three global domains, GL ocean achieved the lowest absolute test error, reaching 0.89 ppm mean annual RMSE over 2021–2025 with a single-hidden-layer MLP of 220 units (22,661 parameters). Whereas previous neural OCO-2 retrieval studies have focused primarily on land nadir or mixed overland soundings (David et al., 2021; Bréon et al., 2022; Xie et al., 2024; Chen et al., 2025), our ocean-glint evaluation establishes a compact baseline for fast multi-year $XCO_2$ estimation over the global oceans.

The lower retrieval RMSE over GL ocean has a clear physical and radiometric basis. Near-glint viewing over water redirects substantially more reflected sunlight toward the sensor than diffuse nadir viewing over dark ocean surfaces (Cox and Munk, 1954; Crisp et al., 2017; O'Dell et al., 2018). Under unsaturated observing conditions, higher photon counts increase the signal-to-noise ratio and reduce relative measurement noise in absorption-line depths, enhancing sensitivity to small $CO_2$-induced spectral variations (Crisp et al., 2017). Because continuum normalization removes broadband radiance scaling without altering relative channel-to-channel noise, this signal-to-noise advantage carries over to the normalized spectra. In addition, the spatially uniform ocean surface reduces lower-boundary variability relative to land, where surface albedo, topography, and continental aerosols vary strongly (Boesch et al., 2011; Massie et al., 2021).

GL ocean also had a smaller prior-error baseline: the 2021–2025 mean annual prior RMSE was 1.74 ppm, compared with 2.67 ppm for GL land and 3.40 ppm for ND (Table 2). MLP reductions of 49 %, 49 % and 58 %, respectively, show comparable proportional gains despite different absolute errors. ACOS represents ocean-glint reflection with a wind-speed-dependent Cox–Munk wave-slope model, whereas land reflection has stronger spatial heterogeneity (Cox and Munk, 1954; Nelson et al., 2020; O'Dell et al., 2018). These radiative and statistical differences provide complementary explanations for the observed domain contrast.

### 4.3 Comparison with previous nadir and regional machine-learning retrievals

The global ND experiment spans diverse observing conditions worldwide and achieved a 2021–2025 mean annual test RMSE of 1.43 ppm. Evaluating the 99-input framework within the East Asian domain (20° N–45° N, 110° E–145° E) places our results in the geographical region examined by Xie et al. (2024) and Chen et al. (2025).

The global ND MLP has 164,793 trainable parameters, compared with approximately 6.5 million estimated from the published SpT architecture (Sect. S5). After aligning the test screening criteria, the frozen MLP produced lower annual RMSEs against ACOS than the published base SpT over the common East Asian test period (Fig. 8a). The result shows that a compact model trained globally can reproduce regional ACOS retrievals without local retraining.

Starting from the same global model, rolling updates improved agreement with ACOS on identical target-quarter soundings. Even without updates, the mean quarterly RMSE gap between the fixed MLP and the published base SpT widened from approximately 0.17 ppm in 2021 to 0.62 ppm in 2023 (Fig. 8b). This pattern points to stronger temporal transfer for the MLP in this East Asian comparison, with rolling updates providing a further reduction in error.

### 4.4 Limitations and future directions

Because our models are trained and evaluated against ACOS Level-2 Standard $XCO_2$, the reported metrics quantify agreement with the operational retrieval rather than accuracy against independent atmospheric observations. The model estimates may inherit systematic biases in the reference product. Evaluating atmospheric accuracy for carbon-flux applications requires independent validation against ground-based Total Carbon Column Observing Network (TCCON) measurements, which serve alongside multi-model ensembles as primary truth-proxies for ACOS v11 evaluation (Taylor et al., 2026). In addition, the 2022 increase in ND prior error coincided with a transition from build B11006r to mixed B11006r and B11008r records, and post-2022 GL-land and ND biases spanned multiple processing builds (Osterman and the OCO-2/OCO-3 Science Team, 2024; Table S11) as well as the operational prior updates and detector-icing period documented in ACOS v11.1–v11.3 (Taylor et al., 2026). Reprocessing matched soundings under a single ACOS build and evaluating against bias-corrected v11.3 products would separate algorithm-version and calibration shifts from atmospheric variability. Future work can also extend this SVD-residual framework to OCO-3, whose v11 $XCO_2$ record agrees with collocated OCO-2 observations to within 0.2 ppm on average (Taylor et al., 2026), and to GOSAT by adapting the wavelength grid and continuum windows to each spectrometer, while using radiative-transfer simulations to isolate how signal-to-noise ratio and higher-order spectral components affect retrieval error.

## 5 Conclusions

We evaluated multi-year OCO-2 $XCO_2$ estimation across global ocean-glint, land-glint, and nadir domains by combining bandwise SVD spectral compression with compact prior-residual MLPs (16,893–164,793 trainable parameters). Trained on 2015–2019 observations and validated on 2020, frozen MLPs generalized across five unseen test years (2021–2025), achieving mean annual RMSEs against ACOS of 0.89, 1.35, and 1.43 ppm for GL ocean, GL land, and ND, respectively (48.7 %–57.9 % reductions relative to the operational prior), while buffering random perturbations to the operational prior and outperforming direct-$XCO_2$ and LightGBM baselines. Under published East Asian ND screening criteria, the frozen global MLP achieved lower annual RMSEs against ACOS than the base SpT in 2020–2022 without regional retraining (mean 1.376 vs. 1.503 ppm), and quarterly rolling updates further lowered the 2021–2024 mean quarterly RMSE from 1.54 to 1.06 ppm. Starting from native spectra and ancillary inputs already in memory, batched CPU preprocessing and inference required 75.6–102.9 μs per sounding across domains. Post-2022 positive biases (+0.45 to +0.65 ppm in 2025) reflect evolving prior residuals and ACOS processing build transitions; independent validation against ground-based TCCON observations remains essential to assess absolute atmospheric accuracy for carbon-flux applications.

## Code and data availability

OCO-2 Level-1B calibrated radiances and ACOS Level-2 Standard $XCO_2$ products (builds B11006r–B11213r) are publicly available from the NASA Goddard Earth Sciences Data and Information Services Center (GES DISC; https://disc.gsfc.nasa.gov/). Analysis code, model configurations, and compact result tables for the 99-input prior-residual framework are publicly available at https://github.com/Sanshi2024/oco2-svd-residual-retrieval.

## Supplement link

The link to the supplement will be included by Copernicus, if applicable.

## Author contributions

XS led the methodology, formal analysis, validation, visualization, and writing of the original draft, and contributed to review and editing. CZ contributed to conceptualization, supervision, funding acquisition, and review and editing. HS contributed to supervision, funding acquisition, and review and editing. KD and SD contributed to review and editing.

## Competing interests

The authors declare that they have no conflict of interest.

## Acknowledgements

We thank the NASA OCO-2 mission and the ACOS science team for providing the calibrated Level-1B spectra and Level-2 retrieval products.

## Financial support

This research has been supported by the Hong Kong Jockey Club Charities Trust (grant no. FA123), the Innovation and Technology Fund (grant no. P0491), the General Research Fund (grant no. 16303725), and the State Key Laboratory of Climate Resilience for Coastal Cities (grant no. ITC-SKLCRCC26EG01).

**Supplement of**

# Compact prior-residual neural networks generalize OCO-2 $XCO_2$ retrievals across viewing modes and five unseen years

Xianlei San[1], Chengxing Zhai[1], Hui Su[1], Kuai Dai[1], and Shuangchen Du[1]

Correspondence to: Chengxing Zhai (cxzhai@ust.hk)

**Sect. S1. Technical details of spectral wavelength alignment, edge trimming, and continuum normalization**

The common wavelength grid and bad-sample statistics were derived from 173,672 valid training granule–footprint combinations, with equal weight assigned to the eight cross-track footprints. Each input file contains 1,016 native samples per band. The trimming rule excluded contiguous edge samples whose bad-sample frequency was at least 30 %, removing 90 leading and 66 trailing samples in $O_2$-A, 90 and 1 in weak $CO_2$, and 90 and 45 in strong $CO_2$. A resampled channel was considered valid only if both bracketing native pixels were unflagged. Continuum normalization used 15 fixed common-grid channels per band selected from high-transmittance baseline segments and required at least three valid samples. The windows were 0.7709826–0.7711585 μm for $O_2$-A, 1.5951169–1.5956051 μm for weak $CO_2$, and 2.0480331–2.0486661 μm for strong $CO_2$ (Fig. S4). SVD bases were fitted without centering on the common-grid normalized training matrices, and per-sounding channel masks were applied during least-squares coefficient estimation; reported variance fractions were calculated from centered score moments in float64.

**Sect. S2. Model selection and optimization settings**

LightGBM minimized squared error on residuals in ppm, using at most 3,000 boosting rounds and validation early-stopping patience of 100. The 24 configurations comprised a predefined reference and 23 unique grid draws with random seed 20261025. The search space comprised learning rates {0.01, 0.03, 0.07}, leaves {15, 31, 63, 127}, minimum leaf populations {20, 100, 300}, feature fractions {0.7, 0.9, 1.0}, bagging fractions {0.8, 1.0}, and L2 penalties {0, 1, 10}; bagging frequency was 1. Seed 17 screened candidates on the complete training and validation cohorts; the three best candidates and the reference were confirmed with seeds 17, 29 and 43. Mean validation RMSE determined the final configuration.

The selected global configurations used learning rate 0.07, 127 leaves, at least 300 samples per leaf, feature fraction 0.9, bagging fraction 0.8 and L2 penalty 10. The three selections were frozen before subsequent-year evaluation. The nine selected fits reached the 3,000-round budget; prediction used the iteration with the best performance on the 2020 validation cohort.

Each MLP minimized minibatch MSE on residuals standardized by their training mean and population standard deviation, using AdamW ($\beta_1 = 0.9$, $\beta_2 = 0.999$, $\varepsilon = 10^{-8}$) and gradient-norm clipping at 5. Weight decay was $10^{-5}$ for global GL ocean, 0 for GL land, $10^{-6}$ for global ND. Learning rates, batch sizes, effective dropout, and best/stopping epochs are reported in Table S7; scheduler and regularization settings are summarized in Table S12.

All regression models used the 99-input representation. Domain-specific MLP nonarchitecture settings were held fixed (Tables S7 and S12) during the hidden-layer architecture search on the corresponding 2020 validation cohort.

The architecture search varied the network depth from one to four hidden layers and the width from 32 to 640 units per layer in steps of four. Optuna TPE used seed 20261018, 24 startup trials, and 64 expected-improvement candidates, with median pruning after 12 startup trials, a 40-epoch grace period, and 10-epoch intervals across a budget of 128 unique completed or pruned candidates. Candidate screening used seed 17 (capped at 200 epochs), and the top three completed architectures in each domain were trained with seeds 17, 29, and 43 and ranked by mean 2020 validation RMSE. Global searches used all 1,000,000 training and 200,000 validation soundings.

Final global models trained for up to 500 epochs (minimum 40 epochs, early-stopping patience 30 epochs, minimum RMSE improvement $10^{-5}$ ppm). The main GL-land model and its component controls both used a 500-epoch cosine horizon. All evaluations used the best validation checkpoint (Table S7). Because dropout was placed between hidden layers, it was active only in the two-layer global ND network. For the ND learning-rate sensitivity check, learning rates of $10^{-4}$, $3 \times 10^{-4}$, $10^{-3}$, and the baseline rate were screened with seed 17 at each component count, and the top two rates were trained across all three seeds and selected by mean 2020 validation RMSE before evaluating the 2021–2025 test sets (Table S10).

**Table S1. Sample counts for the global modelling cohorts and the East Asian ND test cohort.**

| Cohort | Period | Soundings | Role |
|---|---|---|---|
| Global / each domain | 2015-2019 | 1,000,000 | train |
| Global / each domain | 2020 | 200,000 | validation |
| Global / each domain | 2021 | 199,995 | test |
| Global / each domain | 2022 | 200,000 | test |
| Global / each domain | 2023 | 199,997 | test |
| Global / each domain | 2024 | 200,000 | test |
| Global / each domain | 2025 | 195,237 | test |
| East Asia ND | 2020 | 125,897 | test |
| East Asia ND | 2021 | 134,088 | test |
| East Asia ND | 2022 | 87,334 | test |
| East Asia ND | 2023 | 56,383 | test |
| East Asia ND | 2024 | 53,098 | test |
| East Asia ND | 2025 | 18,334 | test |

**Table S2. Global residual-model performance and the operational prior baseline in reconstructed $XCO_2$ space (three-seed means; 2021–2025 RMSE is the unweighted mean of annual test RMSEs; bias is prediction minus reference).**

| Domain | Target | Model | 2020 RMSE | 2021–25 RMSE | 2025 RMSE | 2025 bias | 2025 $R^2$ |
|---|---|---|---|---|---|---|---|
| GL ocean | Prior | — | 1.693 | 1.743 | 1.644 | 0.739 | 0.499 |
| GL ocean | Residual | MLP | 0.648 | 0.894 | 1.069 | 0.452 | 0.786 |
| GL ocean | Residual | LightGBM | 0.865 | 1.424 | 1.951 | 1.514 | 0.293 |
| GL land | Prior | — | 2.490 | 2.665 | 2.514 | 0.870 | 0.287 |
| GL land | Residual | MLP | 1.024 | 1.348 | 1.381 | 0.632 | 0.784 |
| GL land | Residual | LightGBM | 1.266 | 1.724 | 1.990 | 1.139 | 0.553 |
| ND | Prior | — | 2.791 | 3.401 | 3.345 | 1.057 | 0.301 |
| ND | Residual | MLP | 0.961 | 1.433 | 1.640 | 0.651 | 0.832 |
| ND | Residual | LightGBM | 1.375 | 1.810 | 2.001 | -0.032 | 0.750 |

**Table S3. Annual GL-ocean performance in reconstructed $XCO_2$ space. Root mean squared error (RMSE), mean absolute error (MAE), and bias are in ppm; values are means across three seeds.**

| Target | Model | Split | n | RMSE | MAE | Bias | $R^2$ |
|---|---|---|---|---|---|---|---|
| Residual | LightGBM | 2015-2019 train | 1,000,000 | 0.515 | 0.396 | -0.000 | 0.985 |
| Residual | LightGBM | 2020 val | 200,000 | 0.865 | 0.662 | 0.056 | 0.830 |
| Residual | LightGBM | 2021 test | 199,995 | 1.009 | 0.795 | 0.474 | 0.773 |
| Residual | LightGBM | 2022 test | 200,000 | 1.230 | 0.993 | 0.762 | 0.658 |
| Residual | LightGBM | 2023 test | 199,997 | 1.537 | 1.262 | 1.118 | 0.458 |
| Residual | LightGBM | 2024 test | 200,000 | 1.394 | 1.113 | 0.827 | 0.608 |
| Residual | LightGBM | 2025 test | 195,237 | 1.951 | 1.667 | 1.514 | 0.293 |
| Residual | MLP | 2015-2019 train | 1,000,000 | 0.546 | 0.407 | -0.032 | 0.983 |
| Residual | MLP | 2020 val | 200,000 | 0.648 | 0.493 | 0.066 | 0.904 |
| Residual | MLP | 2021 test | 199,995 | 0.756 | 0.600 | 0.393 | 0.873 |
| Residual | MLP | 2022 test | 200,000 | 0.818 | 0.647 | 0.368 | 0.849 |
| Residual | MLP | 2023 test | 199,997 | 0.914 | 0.735 | 0.522 | 0.807 |
| Residual | MLP | 2024 test | 200,000 | 0.912 | 0.713 | 0.235 | 0.832 |
| Residual | MLP | 2025 test | 195,237 | 1.069 | 0.852 | 0.452 | 0.786 |

**Table S4. Annual retrieval performance for GL land in reconstructed $XCO_2$ space (same definitions as Table S3).**

| Target | Model | Split | n | RMSE | MAE | Bias | $R^2$ |
|---|---|---|---|---|---|---|---|
| Residual | LightGBM | 2015-2019 train | 1,000,000 | 0.763 | 0.588 | -0.000 | 0.972 |
| Residual | LightGBM | 2020 val | 200,000 | 1.266 | 0.967 | 0.020 | 0.813 |
| Residual | LightGBM | 2021 test | 199,995 | 1.346 | 1.042 | 0.183 | 0.770 |
| Residual | LightGBM | 2022 test | 200,000 | 1.582 | 1.247 | 0.449 | 0.721 |
| Residual | LightGBM | 2023 test | 199,997 | 1.764 | 1.349 | 0.715 | 0.655 |

| Target | Model | Split | n | RMSE | MAE | Bias | $R^2$ |
|---|---|---|---|---|---|---|---|
| Residual | LightGBM | 2024 test | 200,000 | 1.940 | 1.523 | 0.649 | 0.561 |
| Residual | LightGBM | 2025 test | 195,237 | 1.990 | 1.580 | 1.139 | 0.553 |
| Residual | MLP | 2015-2019 train | 1,000,000 | 0.938 | 0.712 | -0.003 | 0.958 |
| Residual | MLP | 2020 val | 200,000 | 1.024 | 0.779 | 0.108 | 0.877 |
| Residual | MLP | 2021 test | 199,995 | 1.080 | 0.832 | 0.384 | 0.852 |
| Residual | MLP | 2022 test | 200,000 | 1.310 | 1.027 | 0.368 | 0.808 |
| Residual | MLP | 2023 test | 199,997 | 1.497 | 1.151 | 0.744 | 0.751 |
| Residual | MLP | 2024 test | 200,000 | 1.472 | 1.159 | 0.673 | 0.747 |
| Residual | MLP | 2025 test | 195,237 | 1.381 | 1.090 | 0.632 | 0.784 |

**Table S5. Annual retrieval performance for global ND in reconstructed $XCO_2$ space (same definitions as Table S3).**

| Target | Model | Split | n | RMSE | MAE | Bias | $R^2$ |
|---|---|---|---|---|---|---|---|
| Residual | LightGBM | 2015-2019 train | 1,000,000 | 0.798 | 0.615 | 0.000 | 0.973 |
| Residual | LightGBM | 2020 val | 200,000 | 1.375 | 1.057 | -0.287 | 0.804 |
| Residual | LightGBM | 2021 test | 199,995 | 1.500 | 1.158 | -0.235 | 0.787 |
| Residual | LightGBM | 2022 test | 200,000 | 1.761 | 1.364 | 0.164 | 0.787 |
| Residual | LightGBM | 2023 test | 199,997 | 1.809 | 1.410 | 0.251 | 0.780 |
| Residual | LightGBM | 2024 test | 200,000 | 1.978 | 1.538 | -0.007 | 0.741 |
| Residual | LightGBM | 2025 test | 195,237 | 2.001 | 1.529 | -0.032 | 0.750 |
| Residual | MLP | 2015-2019 train | 1,000,000 | 0.755 | 0.568 | -0.088 | 0.976 |
| Residual | MLP | 2020 val | 200,000 | 0.961 | 0.732 | 0.201 | 0.904 |
| Residual | MLP | 2021 test | 199,995 | 1.069 | 0.827 | 0.401 | 0.892 |
| Residual | MLP | 2022 test | 200,000 | 1.350 | 1.057 | 0.577 | 0.875 |
| Residual | MLP | 2023 test | 199,997 | 1.512 | 1.201 | 0.780 | 0.846 |
| Residual | MLP | 2024 test | 200,000 | 1.593 | 1.270 | 0.744 | 0.832 |
| Residual | MLP | 2025 test | 195,237 | 1.640 | 1.311 | 0.651 | 0.832 |

**Table S6. Annual performance of the frozen global ND MLP on the East Asian cohort screened using SpT criteria. Metrics are averaged across three seeds; RMSE and bias are in ppm relative to ACOS.**

| Model | Year | n | RMSE | Bias | $R^2$ |
|---|---|---|---|---|---|
| Global ND MLP | 2020 | 125,897 | 1.287 | -0.073 | 0.868 |
| Global ND MLP | 2021 | 134,088 | 1.351 | +0.113 | 0.824 |
| Global ND MLP | 2022 | 87,334 | 1.488 | +0.281 | 0.798 |
| Global ND MLP | 2023 | 56,383 | 1.537 | +0.470 | 0.756 |
| Global ND MLP | 2024 | 53,098 | 1.476 | +0.271 | 0.753 |
| Global ND MLP | 2025 | 18,334 | 1.546 | -0.498 | 0.749 |

**Table S7. Validation-selected residual MLP architectures and training hyperparameters for the global domains.**

| Cohort | Epochs† | Widths | Params | Activation / norm | Dropout | Learning rate / batch |
|---|---|---|---|---|---|---|
| Global GL ocean | 53 / 117 / 72 | 220 | 22,661 | gelu / layernorm | 0 | 0.00247 / 2,048 |
| Global GL land | 56 / 56 / 58 | 164 | 16,893 | gelu / layernorm | 0 | 0.00229 / 16,384 |
| Global ND | 174 / 152 / 206 | 468-248 | 164,793 | gelu / layernorm | 0.15 | 3.81e-05 / 1,024 |

† Stopping epochs are listed in seed order 17/29/43. Global models used maximum/minimum training lengths of 500/40 epochs and early-stopping patience of 30 epochs. In the same seed order, the best epochs were GL ocean 23 / 87 / 42, GL land 26 / 26 / 28, global ND 144 / 122 / 176. All models predict the prior residual. The main land-glint model and its sensitivity controls share the 500-epoch cosine horizon (Sect. S2).

**Table S8. Feature-to-$XCO_2$ inference times, including input standardization, the forward pass, inverse scaling, and prior addition (Sect. 3.5). Values are medians across three seeds, with 100 single-sounding calls and five repetitions of 200,000-sounding batches. Values refer to the validation-selected LightGBM models and the archived MLP implementations of the reported architectures.**

| Domain | Model | Device | Single ms | 200k s | Batch µs/row |
|---|---|---|---|---|---|
| GL ocean | MLP | CPU | 0.163 | 0.193 | 0.96 |
| GL ocean | MLP | GPU | 0.177 | 0.137 | 0.68 |
| GL ocean | LightGBM | CPU | 0.611 | 26.683 | 133.42 |
| GL land | MLP | CPU | 0.171 | 0.202 | 1.01 |
| GL land | MLP | GPU | 0.197 | 0.145 | 0.72 |
| GL land | LightGBM | CPU | 0.661 | 29.364 | 146.82 |
| ND | MLP | CPU | 0.191 | 0.409 | 2.04 |
| ND | MLP | GPU | 0.236 | 0.157 | 0.78 |
| ND | LightGBM | CPU | 1.079 | 28.084 | 140.42 |

**Table S9. In-memory native-radiance-to-$XCO_2$ inference times for the first 1,024 validation soundings in each domain (seed 17). Measurements used five warmed batch repetitions and 50 single-sounding MLP calls and included wavelength registration, normalization, masked projection, and prediction. LightGBM timings were measured for the validation-selected models in a separate session on the same CPU.**

| Domain | Model | Device | 1,024 rows s | Batch µs/row | Single ms |
|---|---|---|---|---|---|
| GL ocean | MLP | CPU | 0.105 | 102.9 | 1.33 |
| GL ocean | MLP | GPU | 0.103 | 101.0 | 2.12 |
| GL ocean | LightGBM | CPU | 0.235 | 229.2 | — |
| GL land | MLP | CPU | 0.094 | 91.9 | 1.21 |
| GL land | MLP | GPU | 0.095 | 93.0 | 1.56 |
| GL land | LightGBM | CPU | 0.240 | 234.7 | — |
| ND | MLP | CPU | 0.077 | 75.6 | 1.20 |
| ND | MLP | GPU | 0.075 | 73.7 | 1.79 |

| Domain | Model | Device | 1,024 rows s | Batch µs/row | Single ms |
|---|---|---|---|---|---|
| ND | LightGBM | CPU | 0.233 | 227.8 | — |

The timing tables characterize the archived implementations of the reported architectures and preprocessing pipeline. The updated land-glint weights retain the same architecture and input dimensions; they were not separately rebenchmarked.

### Sect. S3. Date-block bootstrap and zonal bias diagnostics

Uncertainty in annual mean bias was estimated using 1,000 within-year date-block bootstrap draws, retaining all soundings and seeds for each sampled date; the 2.5th and 97.5th percentiles define the 95 % confidence interval (Fig. 11). Latitudinal profiles were computed for the complete years 2021–2024 by averaging prediction-minus-reference errors across the three seeds and all soundings within 10° latitude bands containing at least 30 observations (Fig. 12).

**Table S10. Spectral component sensitivity with 1,000,000 training and 200,000 validation soundings per domain. Values are three-seed mean RMSEs (ppm). Fixed denotes the domain-specific model settings held constant across component counts; LR-selected denotes the ND learning-rate check selected using only 2020 validation data. Mean test RMSE is the unweighted mean of annual RMSEs for 2021–2025. Smaller inputs use prefixes of the existing 30-score vector without refitting the projection. Training allowed at most 500 epochs, and no final comparison run reached this limit. Land-glint controls used the common 500-epoch cosine learning-rate horizon described in Sect. S2.**

| Domain | Recipe | SVD scores per band | 2020 RMSE | 2021–25 RMSE |
|---|---|---|---|---|
| GL ocean | Fixed | 0 | 1.274 | 1.473 |
| GL ocean | Fixed | 10 | 0.795 | 1.067 |
| GL ocean | Fixed | 20 | 0.694 | 0.909 |
| GL ocean | Fixed | 30 | 0.648 | 0.894 |
| GL land | Fixed | 0 | 1.830 | 2.344 |
| GL land | Fixed | 10 | 1.180 | 1.750 |
| GL land | Fixed | 20 | 1.063 | 1.420 |
| GL land | Fixed | 30 | 1.024 | 1.348 |
| ND | Fixed | 0 | 1.943 | 3.235 |
| ND | Fixed | 10 | 1.064 | 1.462 |
| ND | Fixed | 20 | 1.001 | 1.322 |
| ND | Fixed | 30 | 0.961 | 1.433 |
| ND | LR-selected | 0 | 1.943 | 3.235 |
| ND | LR-selected | 10 | 1.019 | 1.438 |
| ND | LR-selected | 20 | 0.983 | 1.359 |
| ND | LR-selected | 30 | 0.961 | 1.433 |

**Table S11. Annual granule-matching counts by ACOS processing build for 2015–2025.**

| Year | Build | Matching records |
|---|---|---|
| 2015 | B11006r | 618 |
| 2016 | B11006r | 4491 |
| 2017 | B11006r | 4408 |
| 2018 | B11006r | 5342 |
| 2019 | B11006r | 4179 |
| 2020 | B11006r | 5186 |
| 2021 | B11006r | 5329 |
| 2022 | B11006r | 757 |
| 2022 | B11008r | 4735 |
| 2023 | B11008r | 5200 |
| 2023 | B11205r | 1 |
| 2024 | B11008r | 2156 |
| 2024 | B11205r | 3601 |
| 2024 | B11213r | 882 |
| 2025 | B11213r | 4089 |

## Sect. S4. Cohort selection and provenance

The spectral pipeline applied the matching and quality criteria described in Sect. 2.1. Eligible records required valid Level-1B metadata, a completed granule record, and footprint indices 0–7. Within each year, the initial domain sampling budget was the minimum of 200,000 and the three eligible domain counts. Sounding identifiers were ranked using unsigned 64-bit SplitMix64 keys with seed 20261014 + year + d × 1000003 (where d = 0, 1, 2 denotes ND, GL ocean, and GL land). The lowest-key records were selected without replacement and restored to their original annual row order.

Spectral validity required finite resampled and normalized values across all three bands, more than 50 valid channels per band, and at least three valid channels with positive finite mean radiance in each 15-channel normalization window, together with finite positive reference $XCO_2$ and operational prior values. Surviving counts were equalized across domains using SplitMix64 seed 20261014. The truncated SVD bases were fitted to all 1,000,000 selected 2015–2019 training spectra per domain.

Prior to final projection, duplicate sounding identifiers were screened for consistency across mode, domain, footprint, scalar, normalized-spectrum, continuum, and mask fields, retaining a single copy of agreeing records and rebalancing domain counts with seed 20261014. Rank-deficient masked fits and nonfinite projection scores were excluded, ensuring unique sounding identifiers in each final annual cohort.

The annual projection manifests record the ordered sounding identifiers, archive row indices, domain and mode labels, granule and footprint indices, resampled positions, reference $XCO_2$ and operational priors for 2015–2025.

For East Asian evaluation we used the criteria in Sect. 2.6 on original Level-2 records, joined by sounding ID to the corresponding archived Level-1B spectra. A fixed seed interleaved monthly batches within each year. The combined 2020–2025 test pool was capped at 500,000 unique soundings, with at most approximately 80 % of eligible batches read per year; the cap was not reached. Missing file pairs, invalid required inputs and duplicate soundings were excluded. Removing soundings shared with the original global ND validation cohort affected only 2020. The 2025 records end on 30 September.

**Table S12. Additional optimization settings for the selected residual MLPs.**

| Cohort | Weight decay | Schedule / horizon | Warmup |
|---|---|---|---|
| GL ocean | 1e-5 | Constant | 0 |
| GL land | 0 | Cosine / 500 | 0 |
| Global ND | 1e-6 | Constant | 0 |

Cosine horizons are scheduler T_max after warmup; minimum learning rate is $10^{-6}$. The GL-land main model and component controls both use a 500-epoch horizon. All MLPs use AdamW, gradient-norm clipping at 5 and MSE on training-standardized residuals. Table S7 reports learning rates, batch sizes, effective dropout and epoch counts.

## Sect. S5. East Asian comparison with the published SpT

Figure 8a uses the annual base SpT RMSEs reported by Chen et al. (2025) for 2020–2022 (1.316, 1.526 and 1.666 ppm). For the frozen global MLP, we compute RMSE from the sounding-level differences from ACOS for each seed and year, average across three seeds, and then average the three annual values with equal weights. The comparison uses published SpT values, not a rerun of its checkpoint on our soundings.

The fixed and rolling models shared 300,376 target-quarter soundings. Quarterly RMSEs were averaged equally over 13 quarters after averaging the three seeds. SpT quarterly values were read from Chen et al. (2025, Fig. 3).

For each target quarter, we initialized from the frozen global ND checkpoints and used screened soundings from the preceding 36 calendar months, starting no earlier than January 2020 and excluding the original global validation soundings. A deterministic ID hash capped sampling at 1,000 soundings per month, with approximately 20 % of sampled observation dates reserved for early stopping. The SVD bases and scalers remained fixed. All MLP weights could be updated with AdamW (learning rate and weight decay $10^{-4}$, batch size 1,024, gradient clipping at 5), with a 500-epoch cap, 40-epoch minimum and patience of 30. The original checkpoint was an admissible choice.

Parameter counts refer to the trainable weights and biases of the regression networks. The global ND MLP contains 164,793 trainable parameters, including trainable normalization terms. Based on the architecture specified in Chen et al. (2025, Table 1) and the corresponding forward-model implementation (https://github.com/TaoRen-Rad/SpT; commit 85d1f953beaaba478cff7dfb63a41c42c6f89252), the SpT architecture has an estimated 6,493,953 trainable parameters, including 6,318,080 across the eight Transformer encoder layers and 175,873 across the spectral and ancillary projections, learnable token, positional embeddings, and two-layer 256-unit MLP output head. The five-layer dense network of Xie et al. (2024; 2,564 inputs and hidden widths of 1000, 500, 300, 100, and 20) comprises 3,247,941 trainable parameters. This SpT total is an architecture-based calculation using the published output-head widths, not a count verified from a trained checkpoint. The archived training example instead uses a 128-unit output head; its unused registered encoder template is excluded from the forward-model estimate.

### Sect. S6. Frozen-model perturbations of the $XCO_2$ prior

We retained the 18 adopted residual-model checkpoints, including the validation-selected LightGBM models (three domains, two model families and seeds 17, 29 and 43), the 90 masked-LS30 spectral coefficients, nine ancillary/prior predictors and historical scalers. Each domain used its original 200,000-sounding 2020 and 2024 cohorts. Zero-mean Gaussian noise with standard deviations of 0.5, 1 and 2 ppm was added simultaneously to the $XCO_2$-prior input and the prior added to the predicted residual; all other inputs and ACOS labels were fixed. These evaluations required no training or model selection. Figure 10 shows the 2024 curves, and Table S13 summarizes both years.

For each domain and year, a standardized zero-mean, unit-variance Gaussian noise vector was scaled by $\sigma \in \{0.5, 1, 2\}$ ppm and shared across both model families and all three seeds, with unperturbed predictions serving as the zero-noise baseline. Both perturbed-prior RMSE and model retrieval RMSE were evaluated against the unperturbed ACOS reference and averaged across the three seeds (Table S13).

**Table S13. Sensitivity to zero-mean Gaussian $XCO_2$-prior noise in the original 2020 validation and 2024 test cohorts. Prior and model RMSEs are evaluated against the same ACOS reference; model values are three-seed means. RMSEs are in ppm, and σ is the standard deviation of the added noise in ppm. The prior input and reconstruction prior are perturbed together. Each row represents 200,000 soundings; the baseline uses the unperturbed prior.**

| Domain | Estimate | Year | n | Baseline | σ = 0.5 | σ = 1 | σ = 2 |
|---|---|---|---|---|---|---|---|
| GL ocean | Prior | 2020 | 200,000 | 1.693 | 1.765 | 1.965 | 2.619 |
| GL ocean | Prior | 2024 | 200,000 | 1.580 | 1.658 | 1.871 | 2.551 |
| GL ocean | LightGBM | 2020 | 200,000 | 0.865 | 0.969 | 1.232 | 1.976 |
| GL ocean | LightGBM | 2024 | 200,000 | 1.394 | 1.481 | 1.715 | 2.436 |
| GL ocean | MLP | 2020 | 200,000 | 0.648 | 0.693 | 0.811 | 1.177 |
| GL ocean | MLP | 2024 | 200,000 | 0.912 | 0.955 | 1.075 | 1.461 |
| GL land | Prior | 2020 | 200,000 | 2.490 | 2.538 | 2.680 | 3.188 |
| GL land | Prior | 2024 | 200,000 | 2.734 | 2.780 | 2.913 | 3.390 |
| GL land | LightGBM | 2020 | 200,000 | 1.266 | 1.331 | 1.502 | 2.070 |
| GL land | LightGBM | 2024 | 200,000 | 1.940 | 2.004 | 2.183 | 2.782 |
| GL land | MLP | 2020 | 200,000 | 1.024 | 1.050 | 1.121 | 1.374 |
| GL land | MLP | 2024 | 200,000 | 1.472 | 1.496 | 1.566 | 1.823 |
| ND | Prior | 2020 | 200,000 | 2.791 | 2.836 | 2.966 | 3.436 |
| ND | Prior | 2024 | 200,000 | 3.438 | 3.474 | 3.579 | 3.975 |
| ND | LightGBM | 2020 | 200,000 | 1.375 | 1.424 | 1.552 | 2.032 |
| ND | LightGBM | 2024 | 200,000 | 1.978 | 2.039 | 2.213 | 2.799 |

| Domain | Estimate | Year | n | Baseline | σ = 0.5 | σ = 1 | σ = 2 |
|---|---|---|---|---|---|---|---|
| ND | MLP | 2020 | 200,000 | 0.961 | 0.983 | 1.047 | 1.275 |
| ND | MLP | 2024 | 200,000 | 1.593 | 1.612 | 1.670 | 1.882 |

**Spectral basis vectors for GL ocean**

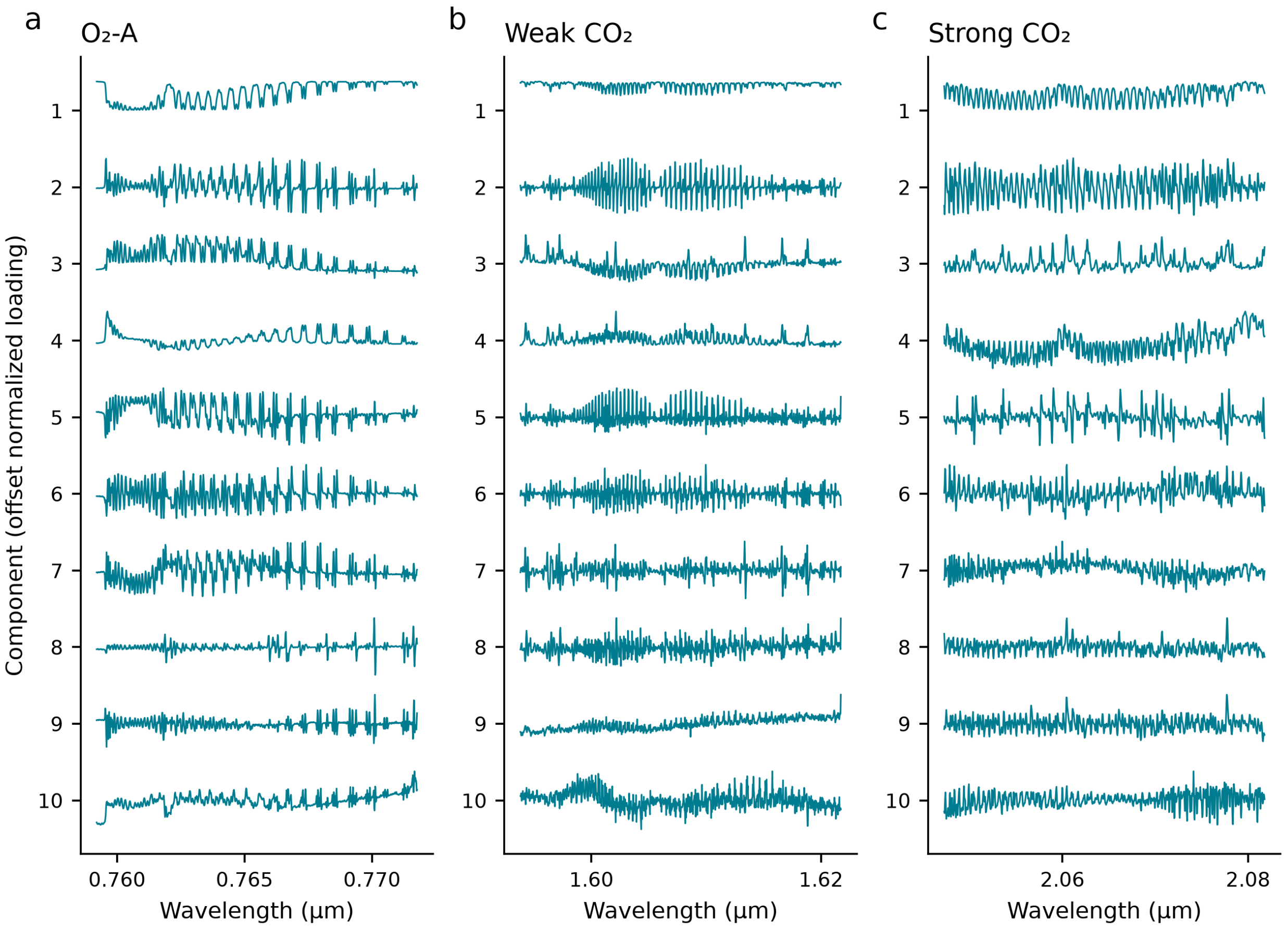

**Figure S1. First ten uncentered SVD right-singular vectors for GL ocean in the (a) $O_2$-A, (b) weak-$CO_2$, and (c) strong-$CO_2$ bands (derived from 2015–2019 training spectra, normalized by maximum absolute loading, and vertically offset for clarity).**

**Spectral basis vectors for GL land**

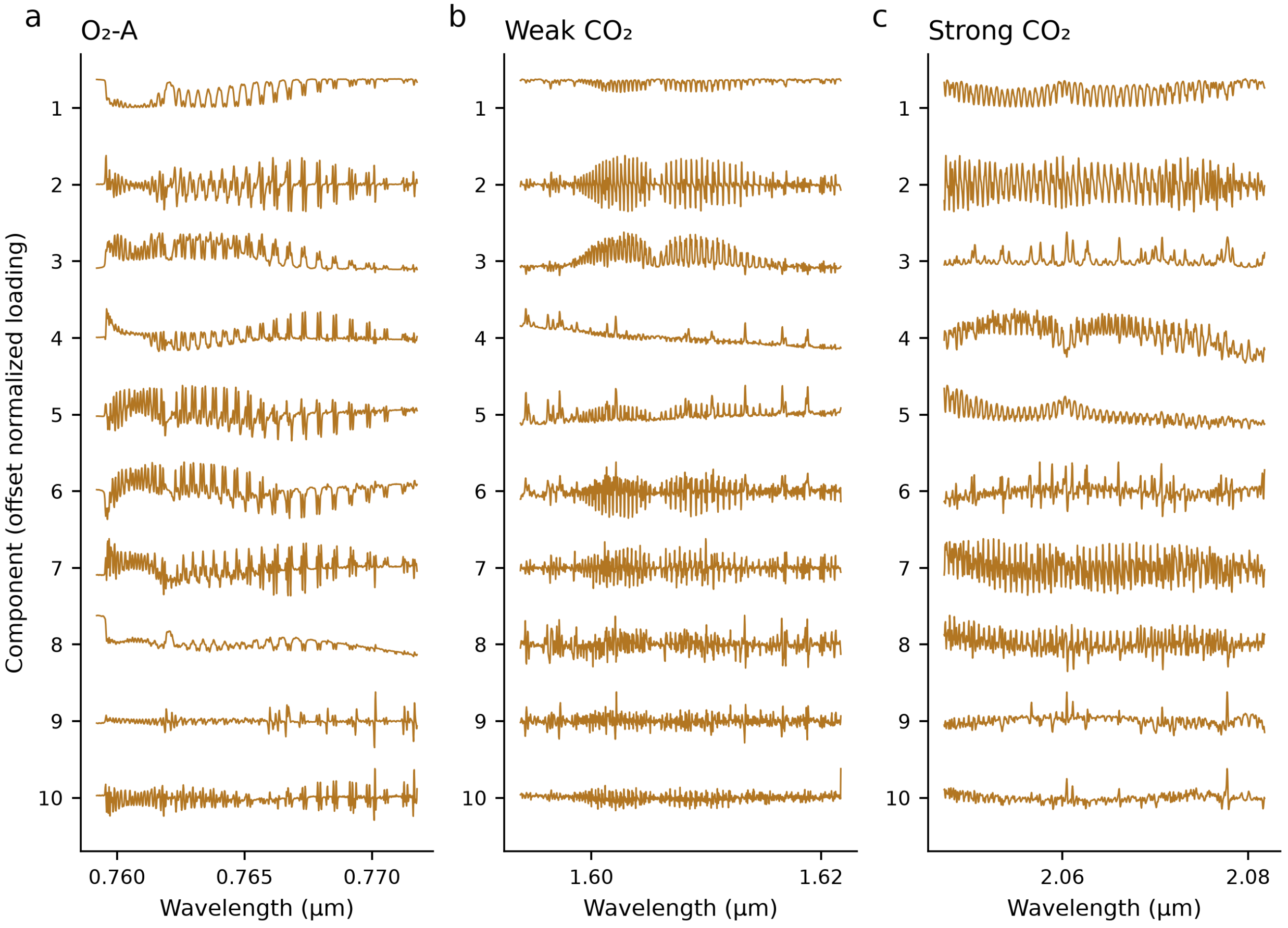

**Figure S2. First ten uncentered SVD right-singular vectors for GL land in the (a) $O_2$-A, (b) weak-$CO_2$, and (c) strong-$CO_2$ bands (normalized and vertically offset as in Fig. S1).**

**Spectral basis vectors for global ND**

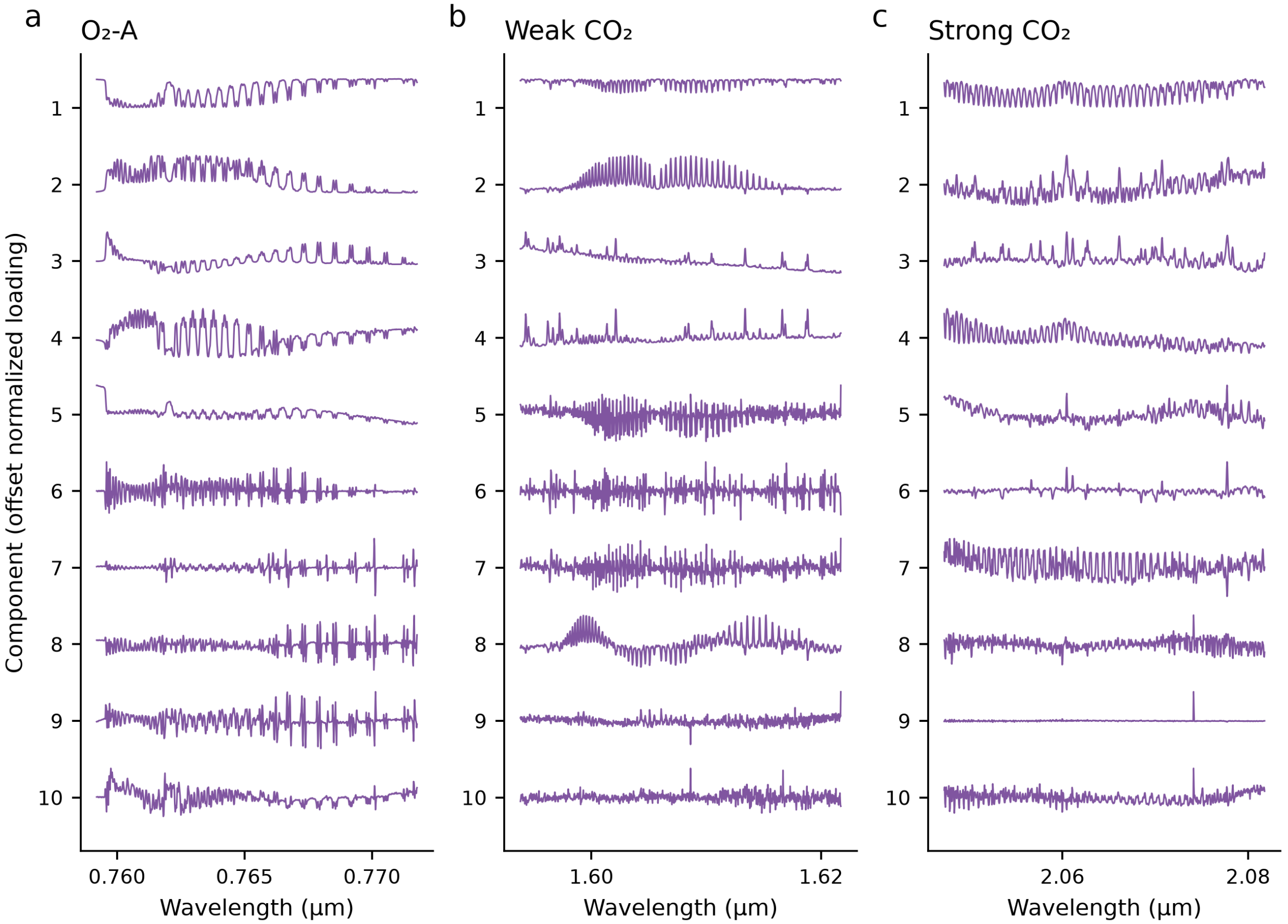

**Figure S3. First ten uncentered SVD right-singular vectors for global ND in the (a) $O_2$-A, (b) weak-$CO_2$, and (c) strong-$CO_2$ bands (normalized and vertically offset as in Fig. S1).**

**Continuum-normalization windows**

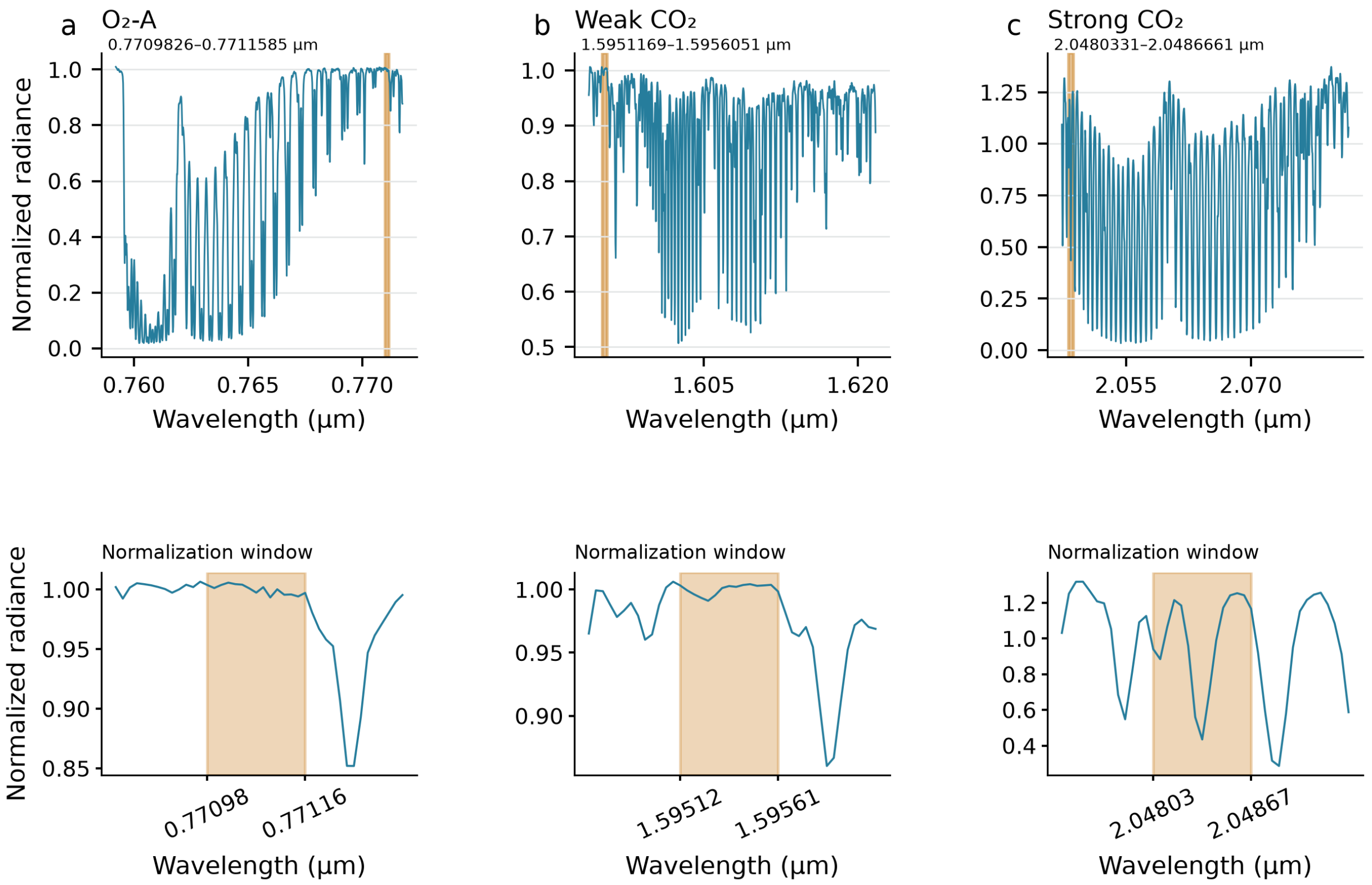

**Figure S4. Fixed continuum-normalization windows on an observed 2020 ocean-glint spectrum. The upper panels show spectra on the retained common grids for $O_2$-A, weak $CO_2$, and strong $CO_2$, with the 15-channel windows shaded; the lower panels enlarge these windows. Each spectrum is normalized using the mean radiance of its valid window channels.**